%% file: main.tex
\documentclass[10pt,twocolumn,letterpaper]{article}

\usepackage{iccv}              

\input{preamble}

\definecolor{iccvblue}{rgb}{0.21,0.49,0.74}
\usepackage[pagebackref,breaklinks,colorlinks,allcolors=iccvblue]{hyperref}

\def\paperID{63} 
\def\confName{ICCV}
\def\confYear{2025}

\title{\ourmethod{}: Human-Aware Multi-view Stereo 3D Reconstruction}

\author{
Sara Rojas\textsuperscript{1,2*} \quad
Matthieu Armando\textsuperscript{2} \quad
Bernard Ghanem\textsuperscript{1} \\[0.2em]
Philippe Weinzaepfel\textsuperscript{2} \quad
Vincent Leroy\textsuperscript{2} \quad
Grégory Rogez\textsuperscript{2} \\[1em]
\textsuperscript{1}KAUST \quad
\textsuperscript{2}NAVER LABS Europe
}

\begin{document}
\twocolumn[{%
	\renewcommand\twocolumn[1][]{#1}
	\maketitle
\vspace{-0.3cm}
    \includegraphics[width=\textwidth]{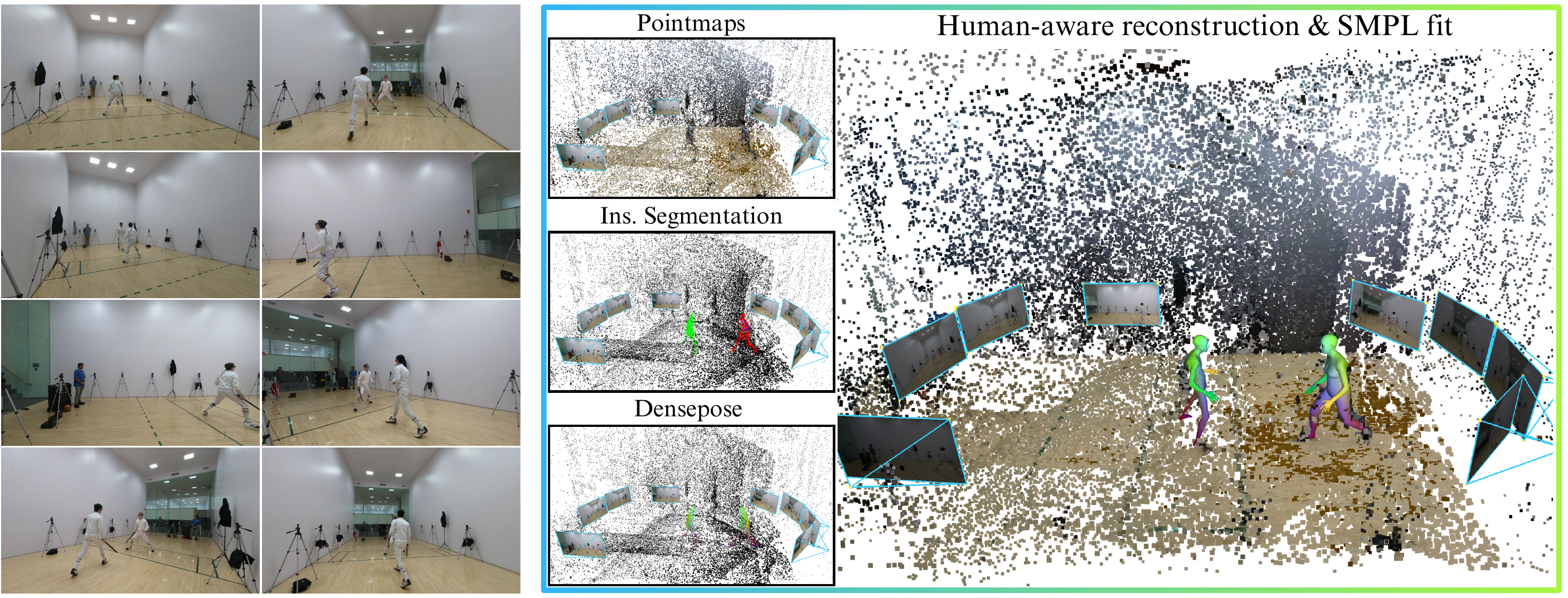} \\[-0.6cm]
    \captionof{figure}{Given a set of unposed images, \ourmethod reconstructs the 3D scene as a dense point map with human semantics, attaching instance segmentation and DensePose information directly to 3D points for human-aware modeling.\label{fig:teaser} \vspace{0.5cm}}
}]
\begingroup
  \renewcommand\thefootnote{\fnsymbol{footnote}}
  \footnotetext[1]{Work done while interning at NAVER LABS Europe.}
\endgroup
\input{sec/0_abstract}    
\input{sec/1_intro}

\input{sec/2_related}
\input{sec/3_method}

\input{sec/4_exps}

\input{sec/5_conclusions}

\paragraph{Acknowledgements.}
The research reported in this publication was partially supported by funding from King Abdullah University of Science and Technology (KAUST) - Center of Excellence for Generative AI, under award number 5940.
\small \bibliographystyle{ieeenat_fullname} \bibliography{main}
\input{sec/10_supp}

\end{document}

%% file: preamble.tex
\usepackage{times}
\usepackage{epsfig}
\usepackage{graphicx}
\usepackage{amsmath}
\usepackage{amssymb}
\usepackage{booktabs}
\usepackage{caption}
\usepackage{color}
\usepackage{xcolor}
\usepackage{multirow}
\usepackage{enumitem}
\usepackage{pifont}
\usepackage[dvipsnames]{xcolor}
\usepackage[ruled, lined,noend, linesnumbered, commentsnumbered]{algorithm2e}
\usepackage{tikz}
\usepackage{xspace}
\usepackage{wrapfig}
\usepackage{comment}
\renewcommand{\paragraph}[1]{\vspace{1mm} \noindent \textbf{#1}}

\def\eg{\emph{e.g}\onedot} 
\def\vs{\emph{v.s}\onedot} 
\def\ie{\emph{i.e}\onedot} 
 
 \def\vs{\emph{vs}\onedot}

\usepackage[accsupp]{axessibility}

\newcommand{\Table}[1]{Table~\ref{tab:#1}}

\newcommand{\ourmethod}{HAMSt3R\xspace}
\newcommand{\divine}{DUNE\xspace}
\newcommand{\master}{MASt3R\xspace}

\definecolor{DarkGreen}{rgb}{0.0, 0.7, 0.5}

\usepackage[normalem]{ulem}
\useunder{\uline}{\ul}{}

\usepackage[precision=2, unit=mm]{lengthconvert}
\usetikzlibrary{patterns}

\usepackage[export]{adjustbox}

%% file: sec/0_abstract.tex
\begin{abstract}

Recovering the 3D geometry of a scene from a sparse set of uncalibrated images is a long-standing problem in computer vision. While recent learning-based approaches such as DUSt3R and MASt3R have demonstrated impressive results by directly predicting dense scene geometry, they are primarily trained on outdoor scenes with static environments and struggle to handle human-centric scenarios. 
In this work, we introduce \ourmethod, an extension of MASt3R for joint human and scene 3D reconstruction from sparse, uncalibrated multi-view images. 
First, we exploit DUNE, a strong image encoder obtained by distilling, among others, the encoders from MASt3R and from a state-of-the-art Human Mesh Recovery (HMR) model, multi-HMR, for a better understanding of scene geometry and human bodies. Our method then incorporates additional network heads to segment people, estimate dense correspondences via DensePose, and predict depth in human-centric environments, enabling a more comprehensive 3D reconstruction. 
By leveraging the outputs of our different heads, \ourmethod  produces a dense point map enriched with human semantic information in 3D.
Unlike existing methods that rely on complex optimization pipelines, our approach is fully feed-forward and efficient, making it suitable for real-world applications.
We evaluate our model on EgoHumans and EgoExo4D, two challenging benchmarks containing diverse human-centric scenarios. Additionally, we validate its generalization to traditional multi-view stereo and multi-view pose regression tasks. Our results demonstrate that our method can reconstruct humans effectively while preserving strong performance in general 3D reconstruction tasks, bridging the gap between human and scene understanding in 3D vision.
\end{abstract}

%% file: sec/1_intro.tex
\vspace{-0.5cm}
\section{Introduction}

3D scene reconstruction from uncalibrated images is a fundamental problem in computer vision with a wide range of applications in robotics, augmented reality, and human-computer interaction. Traditionally, this task has long been approached by solving a succession of problems~\cite{colmapmvs,colmapsfm} using different algorithmic tools like image matching and bundle adjustments. However, recent learning-based methods such as DUSt3R~\cite{dust3r} and MASt3R~\cite{mast3r} introduced a new paradigm by directly regressing the 3D geometry of a scene, given a pair of images. 
These methods not only significantly improved reconstruction quality but also simplified the pipeline, inspiring numerous follow-up works~\cite{cut3r,monst3r,splatt3r}, including some efforts focused on human-centric scene reconstruction~\cite{josh,hsfm}.

Despite these advances, estimating the geometry of scenes involving people remains a major challenge. Humans are highly articulated, exhibit complex deformations, and often appear in self-occluded poses, making their reconstruction significantly more difficult than static environments.
While a complete 3D scene understanding should ideally facilitate Human Mesh Recovery (HMR)—the task of detecting and reconstructing people in 3D—humans themselves provide valuable cues for scene understanding, such as scale estimation.
However, concurrent methods that jointly reconstruct humans and their surrounding environment~\cite{josh,hsfm} rely on cumbersome optimization-based processes, limiting their scalability and practicality.

Furthermore, current learning-based reconstruction models such as MASt3R have been trained on buildings and outdoor scenes, with little focus on human subjects. As a result, they struggle when applied to images containing people, failing to capture articulated structures accurately and often producing incomplete or distorted reconstructions. Addressing this limitation requires integrating additional human-specific cues into the reconstruction pipeline.

In this work, we present \ourmethod which extends MASt3R to explicitly handle human-centric scenes by jointly reconstructing both humans and their surrounding environments (see Figure~\ref{fig:teaser}). To achieve this, we first leverage DUNE~\cite{dune}, a strong image encoder which is pre-trained by distilling those from several teacher models, including MASt3R and Multi-HMR~\cite{multihmr}, a state-of-the-art HMR model, to help the network gain human understanding capabilities.

We then introduce additional processing heads for instance segmentation, dense pose estimation, and binary mask generation. These components allow our model to distinguish human regions from the background, estimate dense correspondences based on the SMPL model~\cite{smpl} (\eg DensePose predictions~\cite{densepose}), and integrate human-specific priors into the reconstruction process. 
By leveraging the outputs of our different heads, \ourmethod  produces a dense point map enriched with human semantic information in 3D. Specifically, each predicted 3D point is classified as human or non-human, with human points mapped to precise body locations of specific individuals. Predictions across multiple images pairs can be aggregated with global alignment, enabling dense, structured human semantics in 3D.

By adapting a state-of-the-art stereo-based reconstruction pipeline to the complexities of human shape recovery, we offer an efficient and scalable alternative to existing optimization-based approaches. Our model effectively bridges the gap between general scene reconstruction and articulated human modeling, enabling high-fidelity 3D reconstructions from sparse and unstructured image collections.
To train the new human-centric heads, we introduce a large-scale, multi-view, synthetic dataset of humans in indoor environments, created by combining the procedural scene generation of Infinigen~\cite{infinigen} with \textit{HumGen3D}~\cite{humgen3d} human generator.

Following~\cite{hsfm}, we evaluate our approach on two challenging human-centric benchmarks, namely EgoHumans~\cite{egohumans} and EgoExo4D~\cite{egoexo}, which feature a variety of indoor and outdoor scenarios with one or several individuals across diverse environments. To ensure that our model maintains strong performance in traditional reconstruction setting - \ie, scenes without humans- we also evaluate it for the task of multi-view stereo depth estimation across several benchmarks following~\cite{dust3r}. Additionally, we assess its ability to perform multi-view pose regression on the CO3Dv2~\cite{co3d} and RealEstate10K~\cite{realestate10K} datasets following~\cite{mast3r}.
Our thorough evaluation shows that our method remains robust across both human-centric and general reconstruction tasks.

The remaining of the paper is organized as follows: after reviewing the related work, we present our methodology, followed by a description of our experiments. Finally, we draw conclusions and discuss potential future directions for improving human-centric 3D scene reconstruction.

%% file: sec/2_related.tex
 \section{Related Work}

We review past work on structure-from-motion, multi-view human reconstruction and both of them jointly.

\paragraph{Structure-from-Motion} (SfM)~\cite{sfmmrf,hybridsfm,mvgeo} consists in reconstructing 3D scene geometry and camera poses given a set of images.
The most popular approach is COLMAP~\cite{colmapsfm,colmapmvs} that relies on traditional feature matching to perform incremental bundle adjustments. For many years, most work has focused on improving various parts of this pipeline such as keypoint detection and description~\cite{sift,surf,orb,r2d2,superpoint}, feature matching~\cite{loftr,superglue,roma}, initialization strategies~\cite{agarwal2010bundle,phototourism} or optimization techniques~\cite{barf,levelS2fm}. Recently, there has been a significant paradigm shift towards fully-learnable approaches~\cite{vggsfm,flowmap,dust3r,mast3rsfm,acezero}. In particular, DUSt3R~\cite{dust3r} has shown outstanding performance in unconstrained 3D reconstruction from as few as 2 images. Their core idea is to regress pointmaps for each image, expressed in the coordinate system of the first image. Several extensions have been since proposed including MASt3R~\cite{mast3r} which  also regresses pixel-aligned dense descriptors, Splatt3r~\cite{splatt3r} which  outputs pixel-aligned parameters for 3D Gaussian splatting~\cite{3dgs}, MONSt3R~\cite{monst3r} which enables handling dynamic objects or MUSt3R~\cite{must3r} and CUT3R~\cite{cut3r} which focus on improving efficiency when processing large image sets.
In this paper, we build upon MASt3R to enable joint 3D reconstruction of humans and scenes from sparse uncalibrated views. While prior methods focus on rigid scene reconstruction, our approach explicitly incorporates human understanding while maintaining strong performance on structures, such as buildings and other man-made elements.

\paragraph{Multi-view Human Reconstruction}
 has been extensively studied, particularly in controlled environments where camera parameters are known~\cite{StarckH07,FurukawaP10, sherf,neuman}. In such settings, multi-view geometry can be leveraged for accurate 3D shape estimation, effectively transforming single- and multi-person reconstruction to a triangulation task~\cite{mvgeo}. When intrinsic and extrinsic camera parameters are available, single-view reconstruction techniques can also be extended to multi-view settings by enforcing geometric consistency. For instance, SMPLify~\cite{SMPLify} has been adapted to estimate 3D human body geometry in a shared coordinate frame, where accuracy is assessed by minimizing the 2D reprojection error of keypoints and silhouettes across multiple views, ensuring a geometrically coherent model~\cite{MV-SMPLify}.
Recent approaches have explored setups with unknown camera poses, employing end-to-end learning to jointly estimate camera parameters and 3D human poses~\cite{YuZXTTKP22,uncalipose}. However, these methods are often limited to single-person scenarios~\cite{YuZXTTKP22} or lack scene context integration~\cite{uncalipose}.

\paragraph{Joint Reconstruction of Scene and Humans}
has been studied in some very recent concurrent methods.
JOSH~\cite{josh} begins with an off-the-shelf scene reconstruction model~\cite{dust3r, mast3r}; the resulting geometry offers valuable contact cues that guide human fitting.  
HSfM~\cite{hsfm} instead assumes accurate 2D human keypoints across views to refine camera poses and, in turn, the surrounding scene.  
SynCHMR~\cite{SynCHMR}, following the SLAHMR~\cite{SLAHMR} insight that human meshes can disambiguate SLAM, stitches together camera-frame HMR, monocular depth, and a human-aware SLAM pipeline before a global optimisation fuses the scene, cameras, and a single actor.  
All three pipelines therefore depend on pre-computed modules and iterative refinement to reconcile them.  
Our method removes these dependencies: in a single forward pass, we jointly predict metric 3D point-maps, dense human semantics, and camera parameters, providing a fully integrated and markedly more efficient solution.

%% file: sec/3_method.tex
\section{Methodology}

\begin{figure*}[ht]
    \centering
  
    \includegraphics[width=\textwidth]{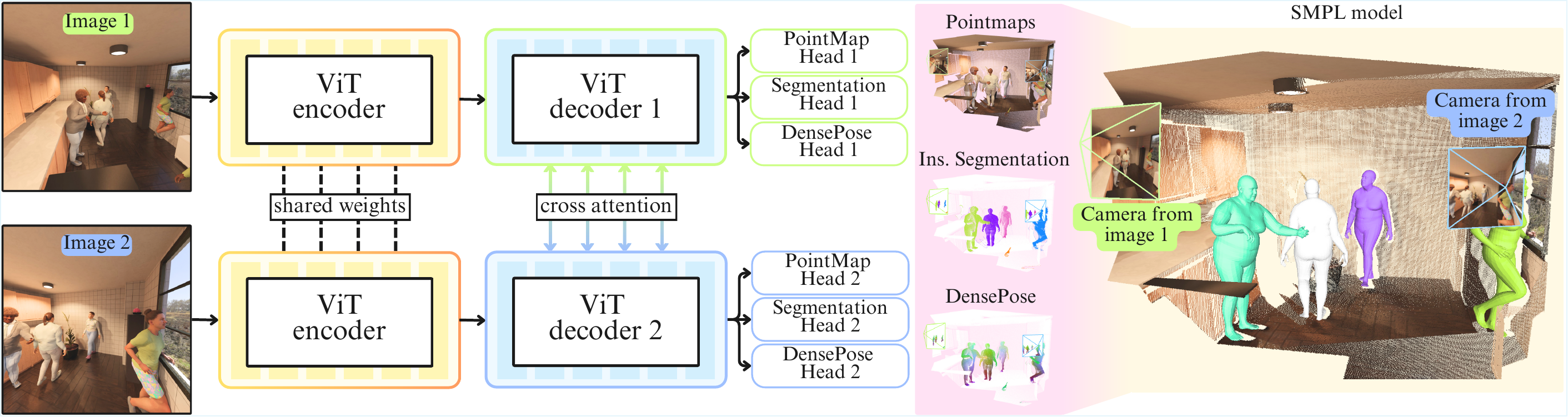}
     \vspace{-0.5cm}
    \caption{\textbf{Overview of \ourmethod.} From left to right: (1) Input stereo images are processed via a Siamese ViT encoder, (2) extracted features are passed to dual decoders with cross-attention, (3) separate heads generate 3D pointmaps and dense human semantic information, in the form of instance segmentation, DensePose, and binary mask predictions. (4) These outputs can be lifted to 3D using the Pointmaps and can be used, for example, to fit a SMPL body model for each human.}
    \label{fig:pipeline_diagram}
    \vspace{-0.4cm}
\end{figure*}

An overview of our model is shown in Figure~\ref{fig:pipeline_diagram}.
Building on the DUSt3R/MASt3R architecture, our approach takes two input  images $I_0$ and $I_1$ which are encoded using a siamese ViT encoder. Each image is then processed by a separate ViT decoder where cross-attention is applied with the tokens from the other image. In addition to the 3D head of MASt3R (which predicts pointmaps and descriptor features for matching), our model is also trained to produce additional human-specific outputs, including human instance segmentation and DensePose predictions~\cite{densepose}. 
This helps the model gain a more meaningful understanding of the human geometry and its relation to the scene, adding rich semantic information to the reconstruction process.
These outputs can be further leveraged for tasks like SMPL predictions through an optimization procedure. We first provide background on \master (Section~\ref{sub:background}), then describe the image encoder adopted from~\cite{dune} (Section~\ref{sub:encoder}) and the additional heads and training strategy (Section~\ref{sub:method}). 

\subsection{Background on MASt3R}
\label{sub:background}

Given an image pair $I_0,I_1 \in \mathbb{R}^{H \times W \times 3}$ (3 channels for RGB), \master~\cite{mast3r} jointly performs  local 3D reconstruction and pixelwise matching. To achieve this, the network predicts a pixel-aligned pointmap $X \in \mathbb{R}^{H \times W \times 3}$ for each image, \ie,   containing the predicted 3D coordinates of the scene point corresponding to each pixel, expressed in the coordinate system of the first image's camera. A confidence map $C$ is also produced. Additionally, another head predicts a small descriptor for each pixel, enabling efficient matching via approximate but fast nearest neighbor search. 

For training, \master uses a confidence-aware regression loss for pointmaps and a InfoNCE loss for the local descriptor learning. We denote its total loss as $\mathcal{L}_{\text{MASt3R}}$.

In terms of architecture, each image is processed by a ViT encoder $Enc$~\cite{vit} to obtain a feature map $F = Enc(I) \in \mathbb{R}^{h \times w \times d}$ for an image $I$.
A dual ViT decoder $Dec$, incorporating cross-attention blocks, then processes both feature maps while attending to the tokens from the other image: $F'_0, F'_1 = Dec \big( Enc(I_0), Enc(I_1) \big)$.
Finally, the prediction heads operate on $F'_0$ or $F'_1$, producing pixelwise outputs via either a linear head or a DPT head~\cite{dpt}.
In this paper, we exclusively use linear heads.

\subsection{A Strong Image Encoder}
\label{sub:encoder}
We replace the original MASt3R encoder with a stronger image encoder $Enc$: a distilled ViT-B/14 backbone obtained through the multi-teacher strategy of DUNE~\cite{dune}. Distillation fuses complementary competencies from three powerful teachers—(1) a generalist image encoder (DINOv2~\cite{dinov2}), (2) the encoder of a state-of-the-art multi-person human-mesh-recovery model (Multi-HMR~\cite{multihmr}), and (3) the MASt3R encoder itself~\cite{mast3r}. Their representations are aligned by the UNIC projection mechanism~\cite{unic}, yielding visual features that are simultaneously robust for scenes and humans. Unlike DUNE, which attaches separate decoders per task and therefore reconstructs humans only from single images, our method couples this encoder with a unified, end-to-end architecture that jointly performs 3D scene reconstruction, instance segmentation, and cross-view human reconstruction. 

\subsection{\ourmethod}
\label{sub:method}

Unlike \master, which primarily focuses on objects and scenes (\eg indoor environments or buildings), our method --- as shown in Figure~\ref{fig:pipeline_diagram} --- is trained on scenes containing humans. 

\ourmethod takes two images, \(I_0\) and \(I_1\), as input and encodes them using a shared Vision Transformer (ViT) and a cross-attention decoder, generating global feature maps \(F_0\) and \(F_1\). These feature embeddings are then utilized by various heads. In addition to the original point and matching heads, we introduce an object segmentation head that produces segmentation masks for each individual in both images and a DensePose head that predicts DensePose maps for each person, mapping human pixels to the 3D surface of the human body, represented by the SMPL mesh~\cite{smpl}. The following paragraphs describe these additional heads in detail.

\paragraph{Instance Segmentation Head.} 
An instance segmentation head is added to \master, extending the original backbone with a transformer-based design inspired by Mask2Former~\cite{masktwoformer}. This head is specifically designed to segment human instances from the background, generating masks that capture the full appearance of people, including hair and clothing. The segmentation branch is supervised by classification and mask losses following the strategy proposed in~\cite{maskformer,masktwoformer}, but with an extension to account for the two input views. In particular, the classification loss distinguishes between human and background, while the mask loss combines binary cross-entropy and dice loss.  The key idea is that the model's understanding of 3D geometry allows it to assign consistent instance labels to each person across different viewpoints.

\paragraph{DensePose Head.} The DensePose head is introduced as an additional branch to predict DensePose maps using SMPL projection templates. It consists of a linear layer that generates a four-channel prediction: an RGB dense pose map \(P_{\text{dp}} \in \mathbb{R}^{H \times W \times 3}\), where each pixel is assigned an RGB color that encodes its corresponding 3D location on the SMPL template mesh, and a binary mask indicating the regions where the DensePose mapping is valid. 
The DensePose representation, 
when integrated with the 3D reconstruction, 
is expected to enhance the model’s ability to reason about human poses, body parts, and their interaction with the environment, which adds semantics beyond just the raw geometric points.
The predicted DensePose map is supervised by an L2 loss computed as \(\mathcal{L}_{\text{dp}} = \|P_{\text{dp}} - P_{\text{gt}}\|_2^2\), where \(P_{\text{gt}}\) represents the ground truth DensePose map. Unlike the original DensePose approach that uses discrete body-part classes, our method employs a continuous mapping that directly formulates the problem as a 3D regression task, as done in~\cite{crocoman}.

 In addition to the RGB DensePose map, the head produces a binary mask \(M \in \mathbb{R}^{H \times W}\) that specifically distinguishes human regions corresponding to the SMPL projection — this excludes areas such as hair and clothing — from the background. This is in contrast to the instance segmentation mask, which covers the entire human silhouette,  including hair and clothing. The binary mask is optimized using a cross entropy-loss \(\mathcal{L}_{\text{mask}}\), which ensures that the SMPL model can be accurately fitted to the relevant human regions.

\paragraph{Semantic 3D Human Representation.} By combining the outputs of our different heads, we obtain dense point maps with human semantic information in 3D. 
Specifically, every 3D point predicted by our method can be classified as either human or non-human, with human points mapped to specific locations on the body of the corresponding human instance. 
  
An advantage of our method is that we can readily fit the SMPL body model to every detected human, \eg for numerical evaluation (see Section~\ref{sub:xpeval}). This kind of semantic understanding is crucial for applications like tracking, behavior analysis, or interaction with the environment.

\begin{figure}
    \centering
    \includegraphics[width=0.24\linewidth]{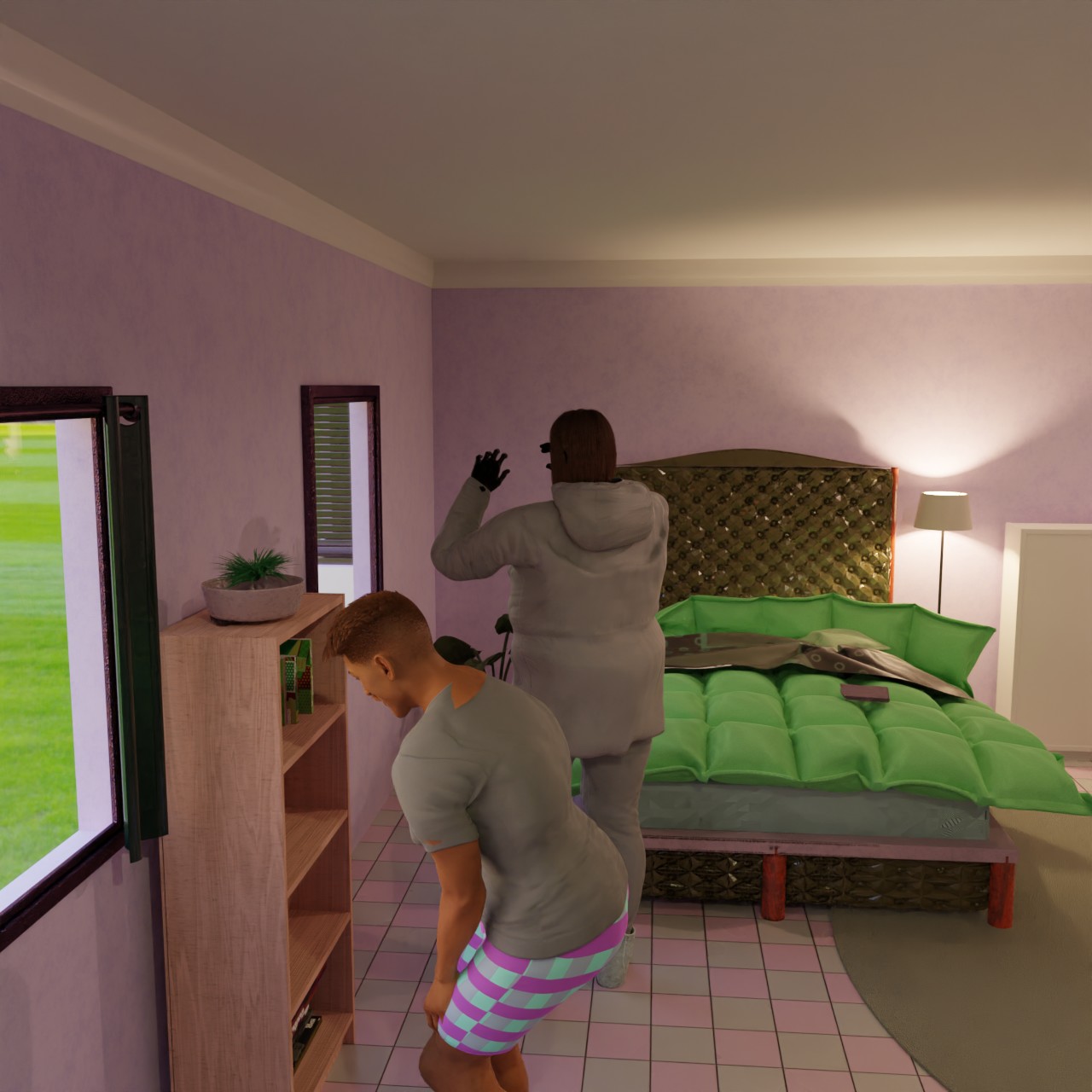}
    \includegraphics[width=0.24\linewidth]{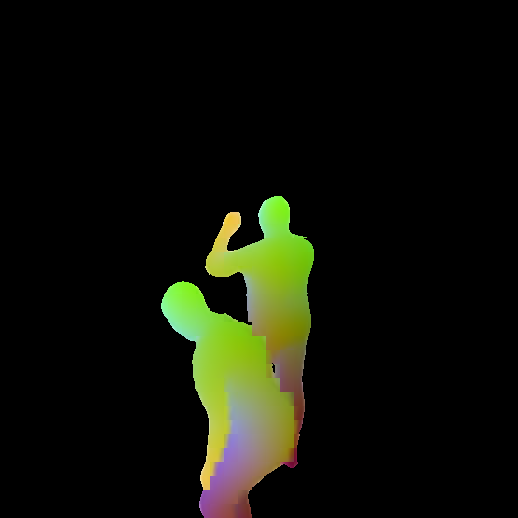}
    \includegraphics[width=0.24\linewidth]{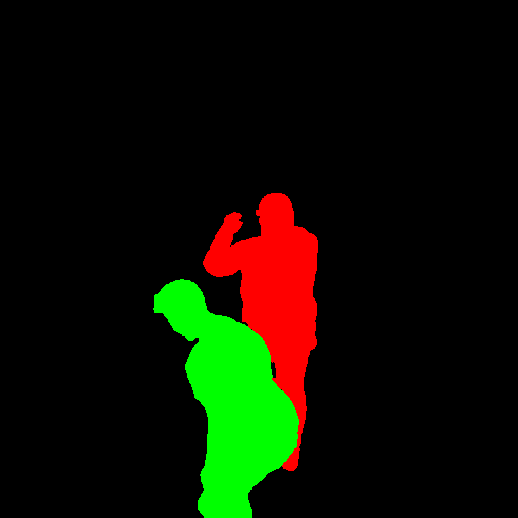}
    \includegraphics[width=0.24\linewidth]{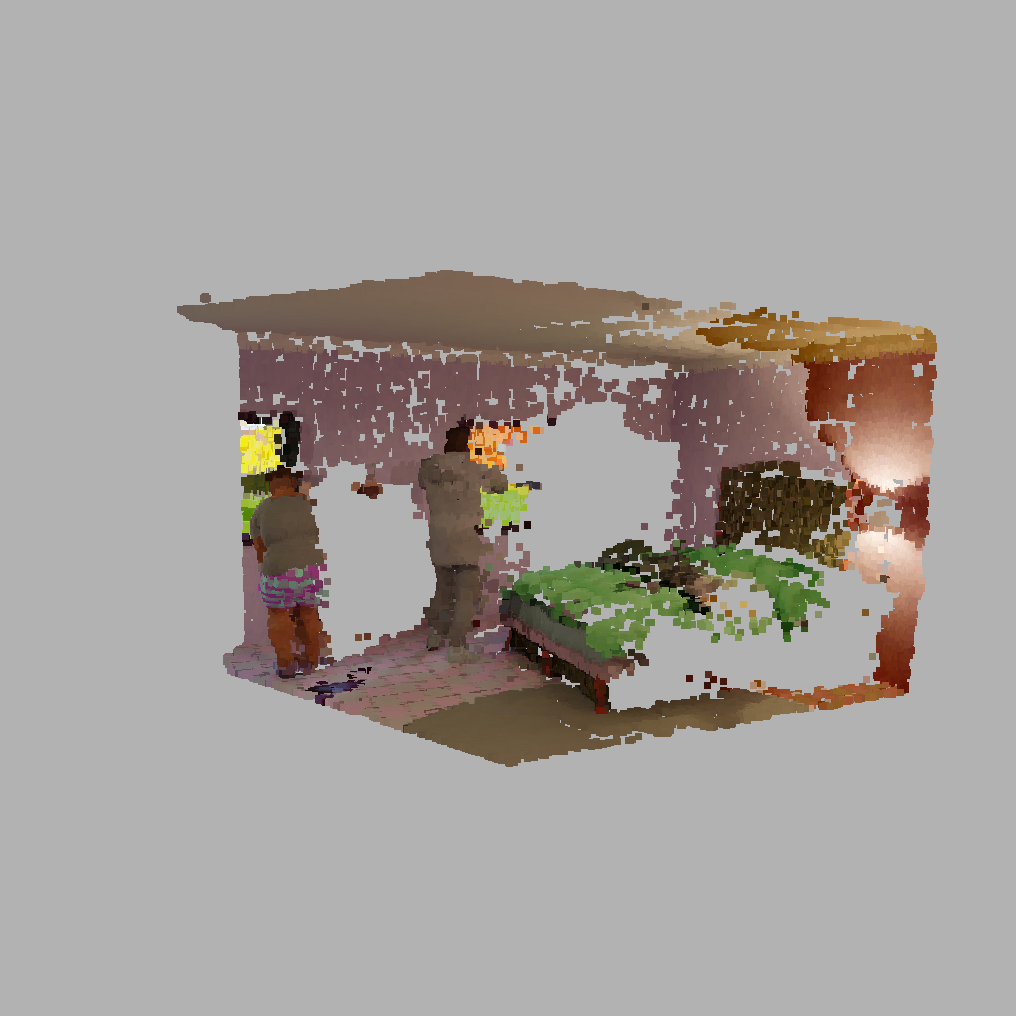} \\
    \includegraphics[width=0.24\linewidth]{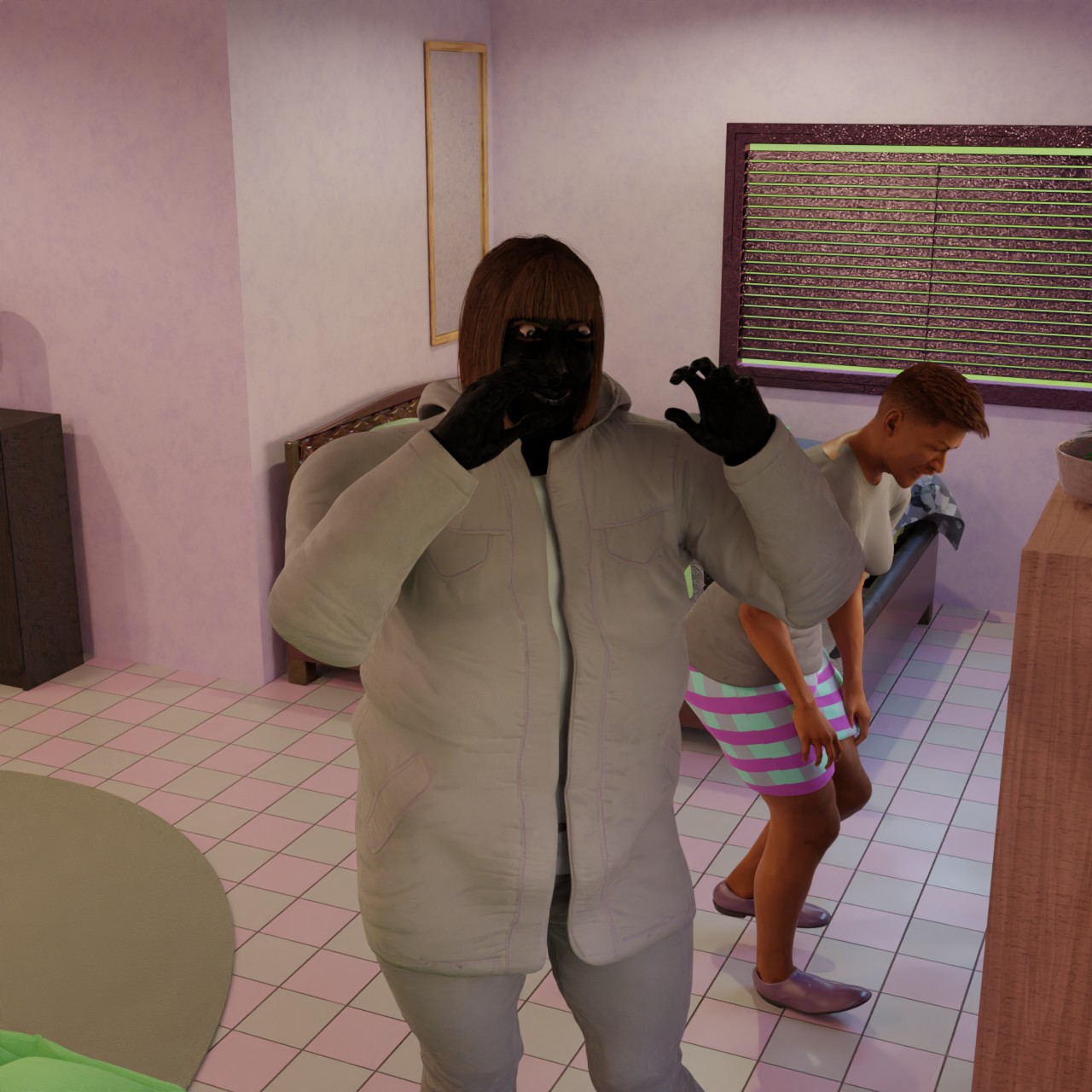}
    \includegraphics[width=0.24\linewidth]{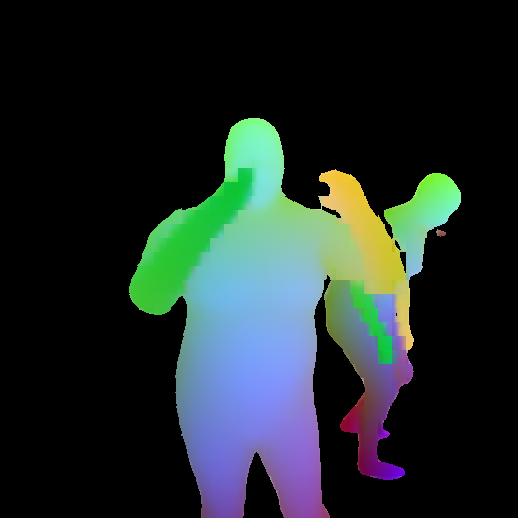}
    \includegraphics[width=0.24\linewidth]{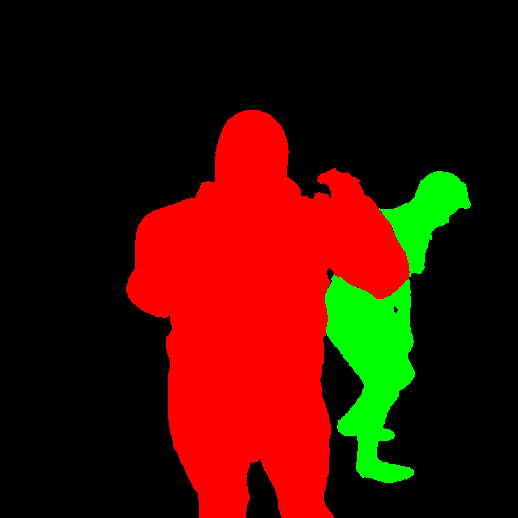}
    \includegraphics[width=0.24\linewidth]{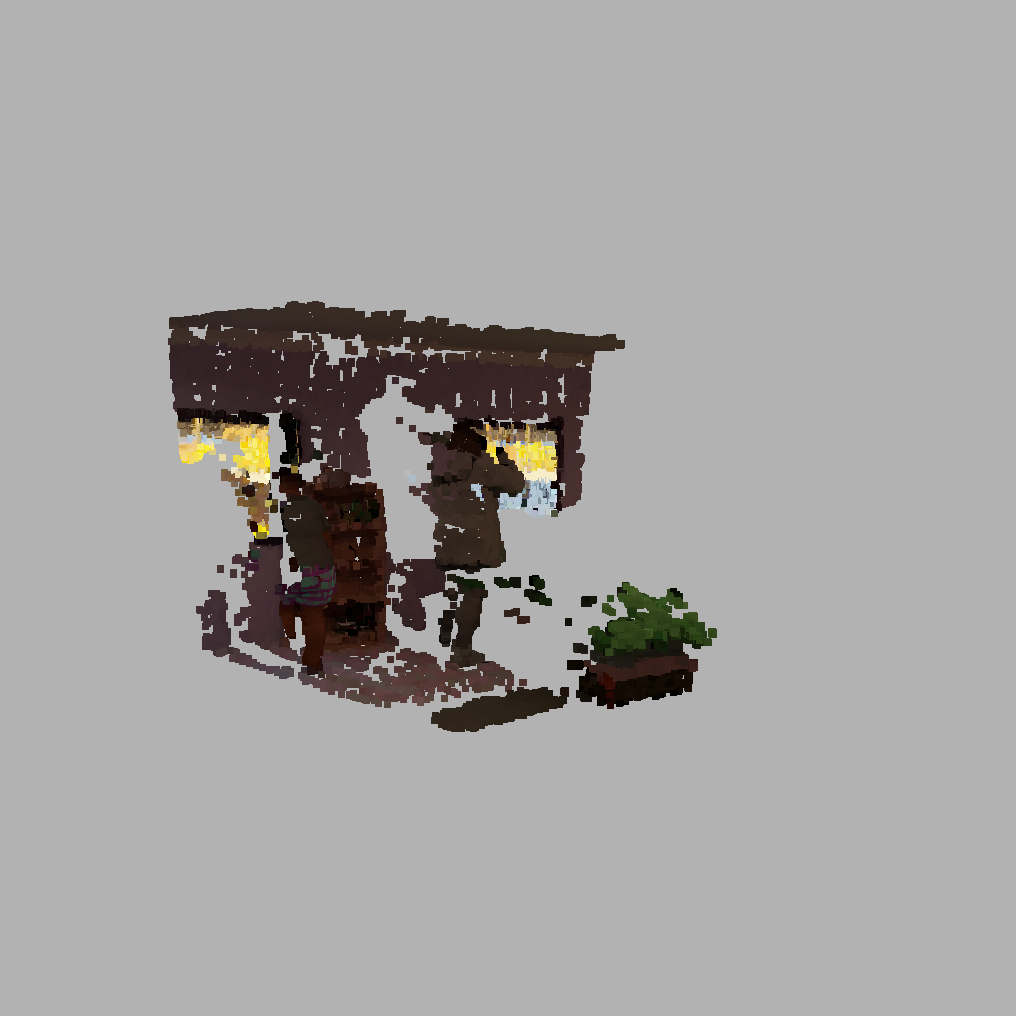} \\
    \includegraphics[width=0.24\linewidth]{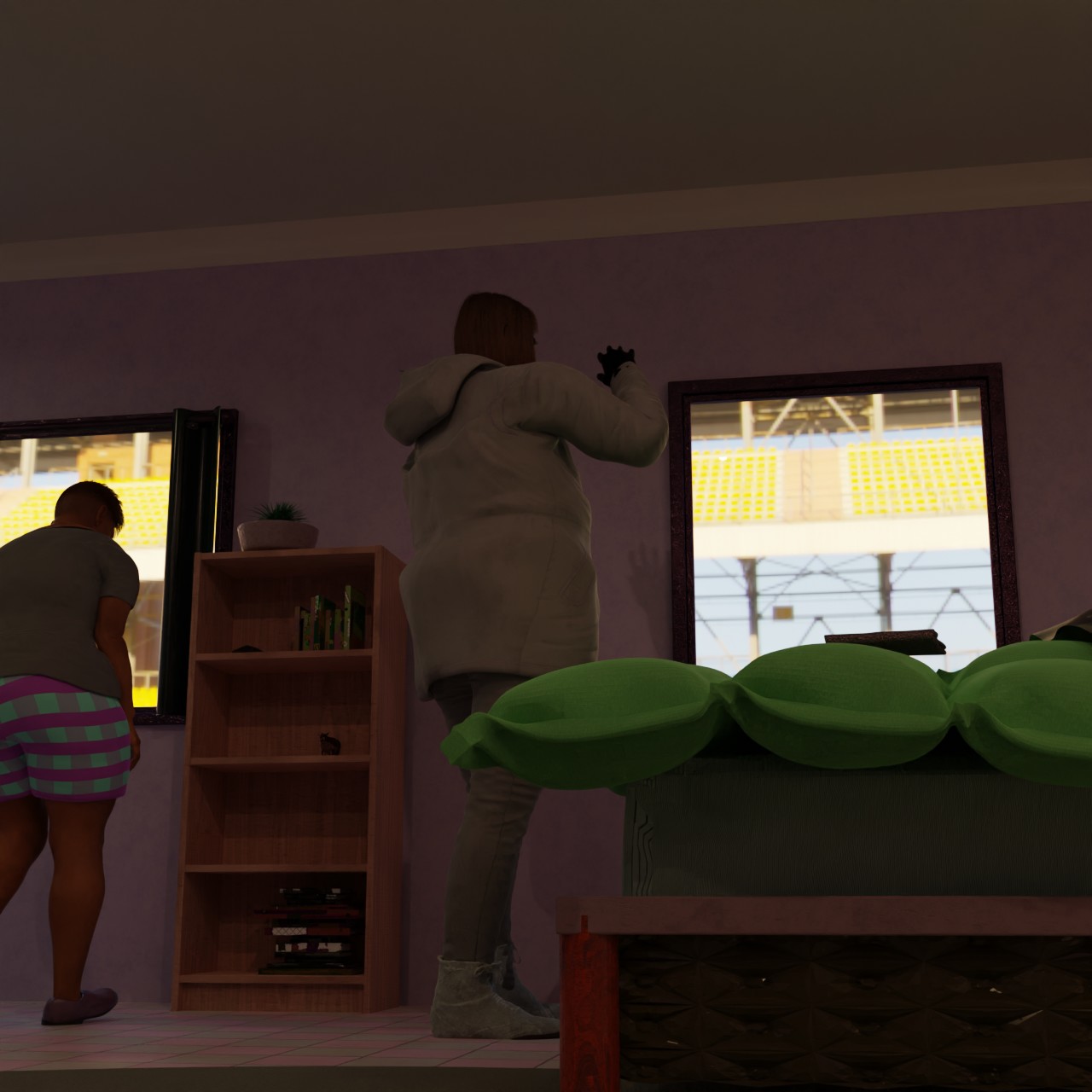}
    \includegraphics[width=0.24\linewidth]{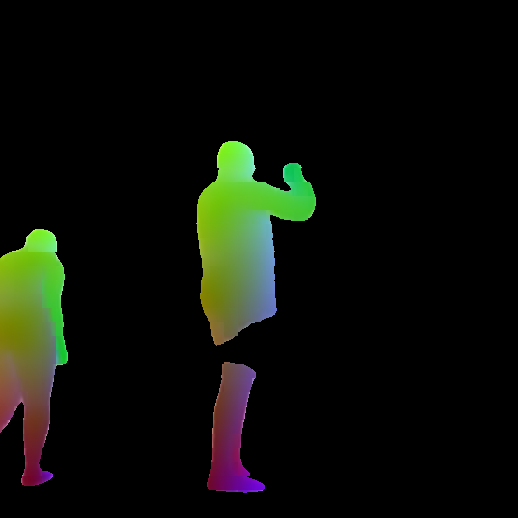}
    \includegraphics[width=0.24\linewidth]{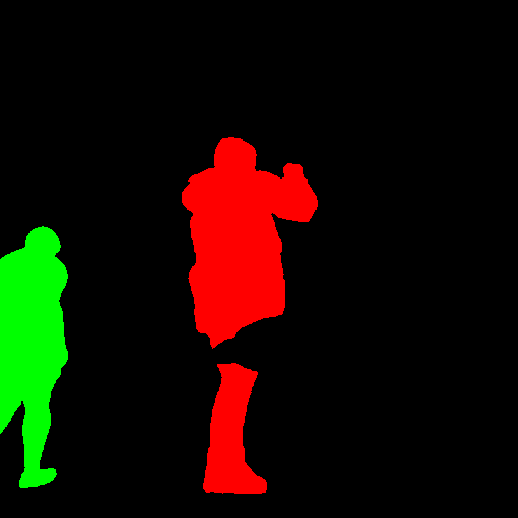}
    \includegraphics[width=0.24\linewidth]{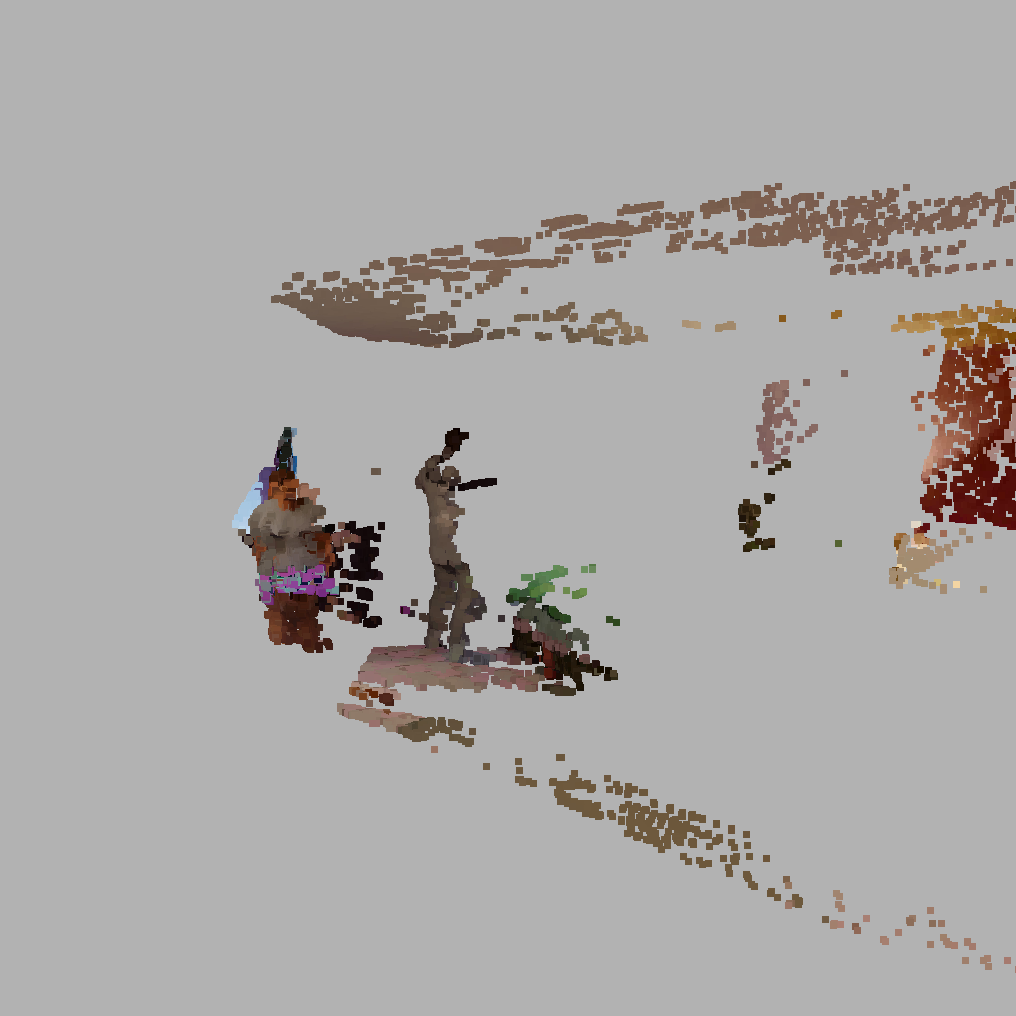} \\

\includegraphics[width=0.99\linewidth]{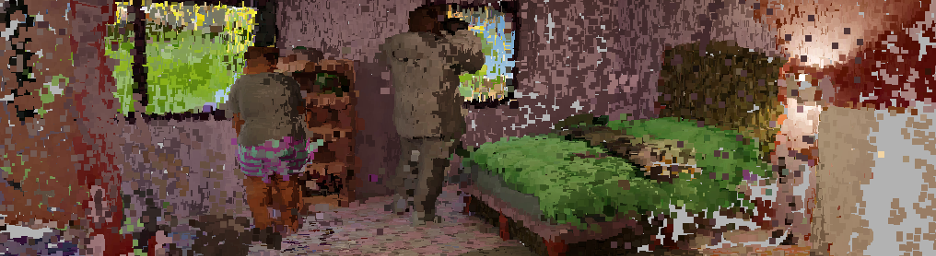}

    \vspace{-0.2cm}
    \caption{\textbf{Results on HumGen3D data} (on a scene not seen during training), using global alignment: Given a set of images of a scene (three out of eight of them are shown in the first column), we run our model on all possible image pairs, and aggregate predictions from the human heads in 2D, for each view (second and third columns). We apply the alignment method of \master~\cite{mast3r} to align the individual pointmaps in 3D (fourth column) and they can be combined into a unified reconstruction (bottom).}
   
    \label{fig:ga}
    \vspace{-0.6cm}
\end{figure}

\paragraph{Dealing with more than 2 views.} 
To handle scenes with an arbitrary number of images, we run our network independently on each possible image pair and align the output pointmaps in 3D using the procedure from \master~\cite{mast3r}. 

Since instance segmentation is performed independently for each pair, person IDs can vary across pairs—for instance, a person labeled as ‘human 1’ in one pair might be labeled as ‘human 2’ in another. To maintain consistent IDs across the scene, we resolve ID correspondences using 2D overlap before integrating all pairs into a consistent 3D representation.
For DensePose, we aggregate the predictions for a same image (produced by each pair) by performing a weighted average of the DensePose outputs, using confidence scores as weights. This prioritizes higher-confidence predictions, resulting in a more accurate and stable human surface representation in 3D.
Predictions from multiple image pairs are subsequently combined after a global alignment step, ensuring a coherent 3D representation of human semantics across the entire scene. This is illustrated in Figure~\ref{fig:ga}.
Note that in the binocular case, our approach is fully feed-forward.

\paragraph{Training.} 
The \divine image encoder is frozen to preserve the distilled features while the decoders and the new heads are fine-tuned. The overall training loss $\mathcal{L}$ is a weighted sum of the MASt3R loss \( \mathcal{L}_{\text{MASt3R}} \), the segmentation loss \( \mathcal{L}_{\text{seg}} \), the dense pose loss \( \mathcal{L}_{\text{dp}} \), and the binary mask loss \( \mathcal{L}_{\text{mask}} \):
\begin{equation}
\mathcal{L} = \mathcal{L}_{\text{MASt3R}} + \lambda_1 \mathcal{L}_{\text{seg}} + \lambda_2 \mathcal{L}_{\text{dp}} + \lambda_3 \mathcal{L}_{\text{mask}},
\end{equation}
where $\lambda_1$, $\lambda_2$ and $\lambda_3$ are loss weights; details on their selection can be found in the Supp. Mat.
Training is performed by mixing 50\% of the original \master dataset with 50\% human-specific data in each epoch. For the original \master dataset, supervision is exclusively applied to the point maps and the matching head. In contrast, the human-specific data  is used to supervise all heads. This tailored supervision strategy ensures that each network component is optimally trained based on the available data. 
Training samples consist of image pairs from multi-camera setups or closely spaced frames from videos, ensuring diverse viewpoints while maintaining spatial coherence. All images are downscaled to a maximum dimension of 518 pixels.

\input{tables/datasets}

\begin{figure}
    \centering
    \includegraphics[width=0.49\linewidth]{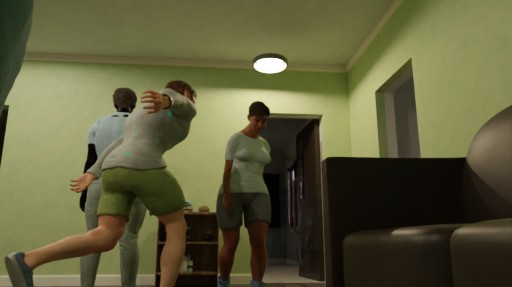}
    \includegraphics[width=0.49\linewidth]{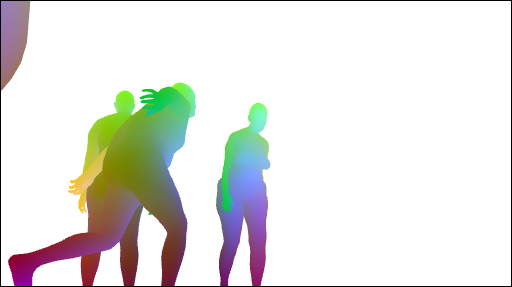} \\\includegraphics[width=0.49\linewidth]{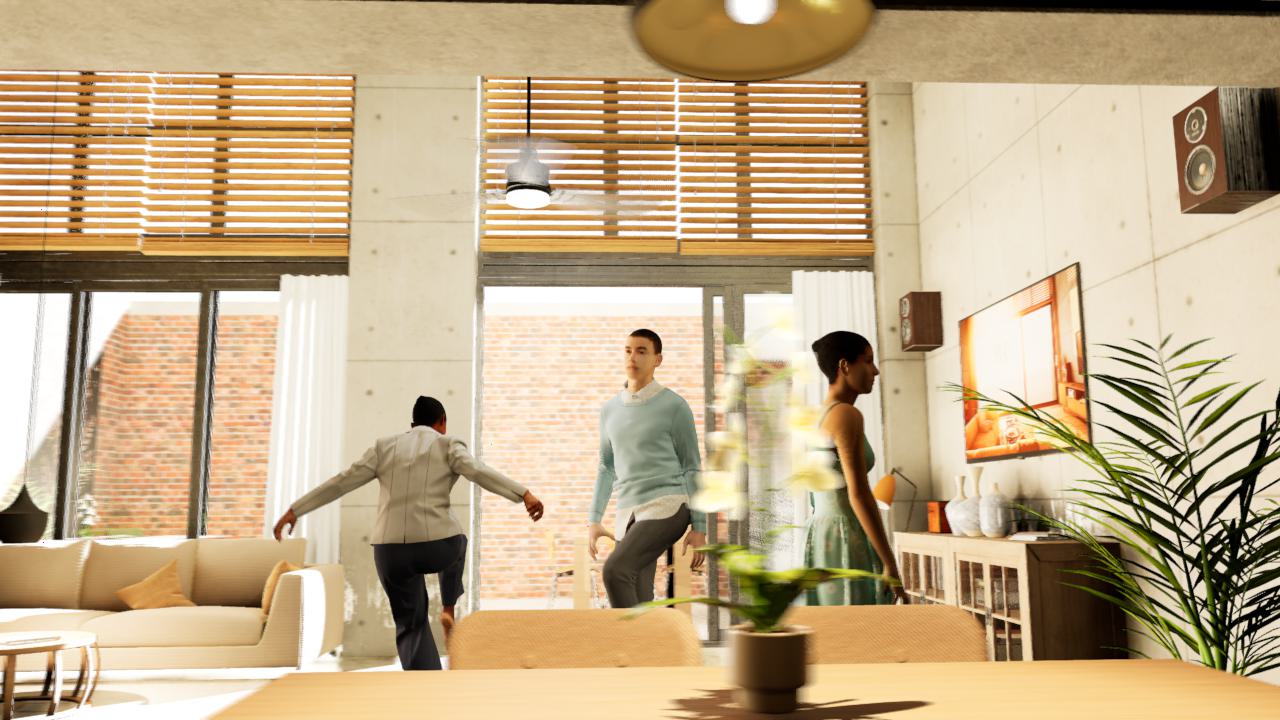}
    \includegraphics[width=0.49\linewidth]{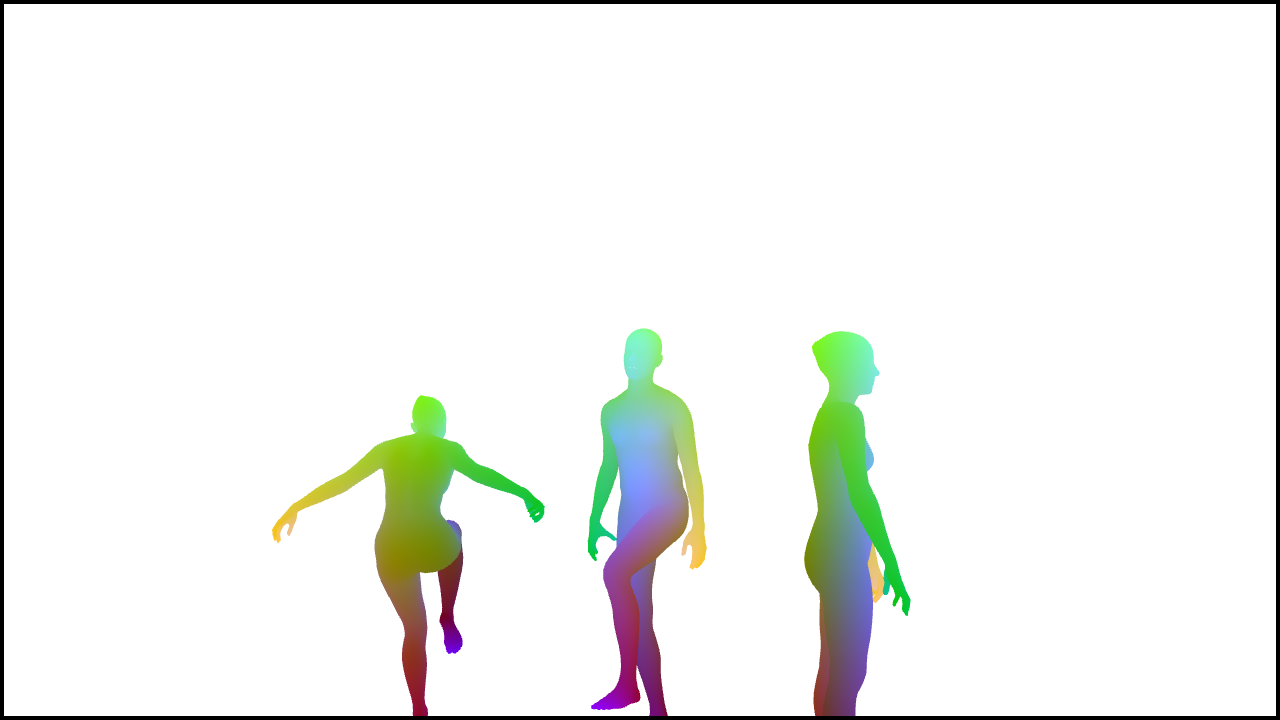} \\
    \includegraphics[width=0.49\linewidth]{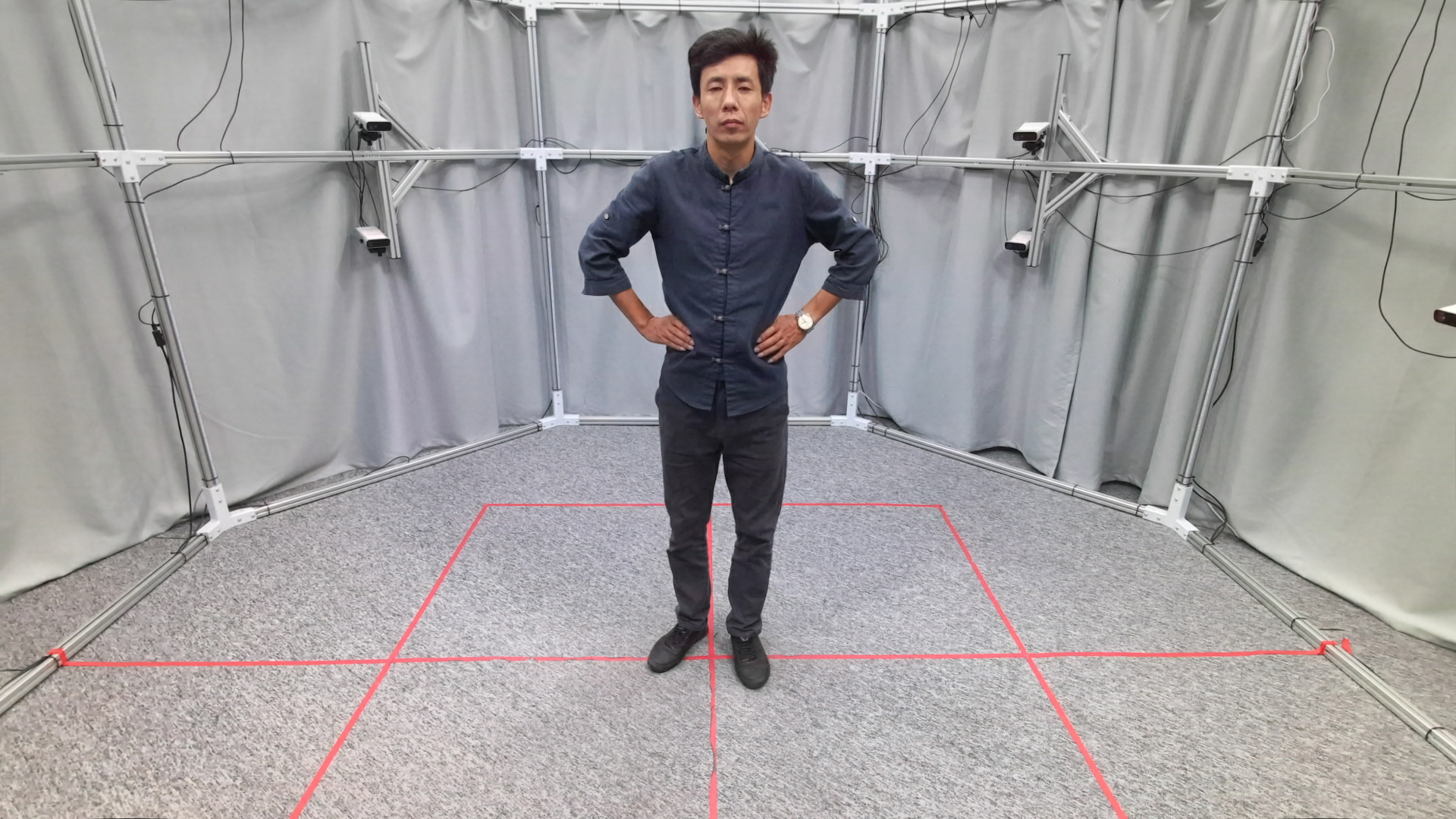}
    \includegraphics[width=0.49\linewidth]{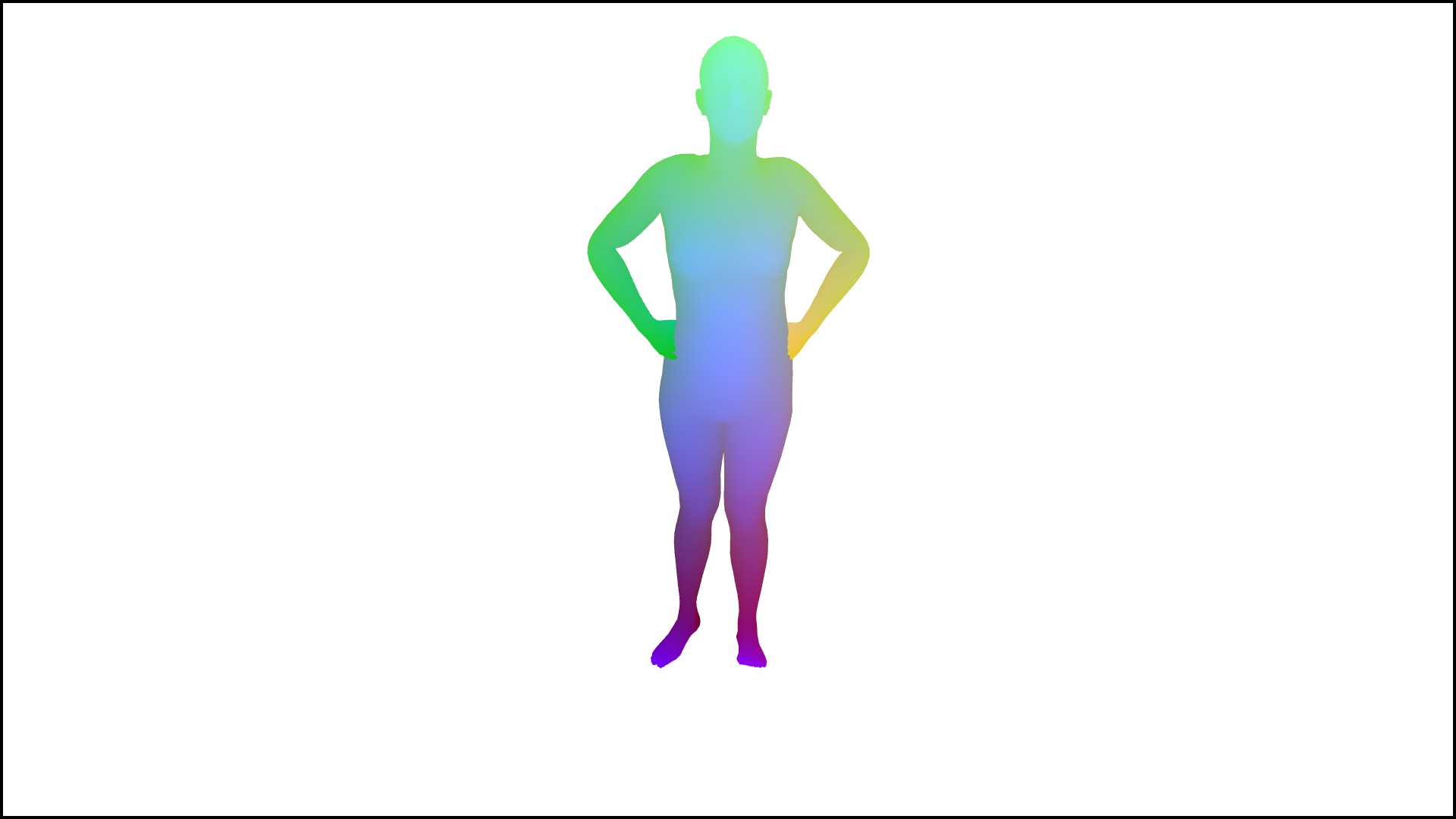} \
    \includegraphics[width=0.49\linewidth]{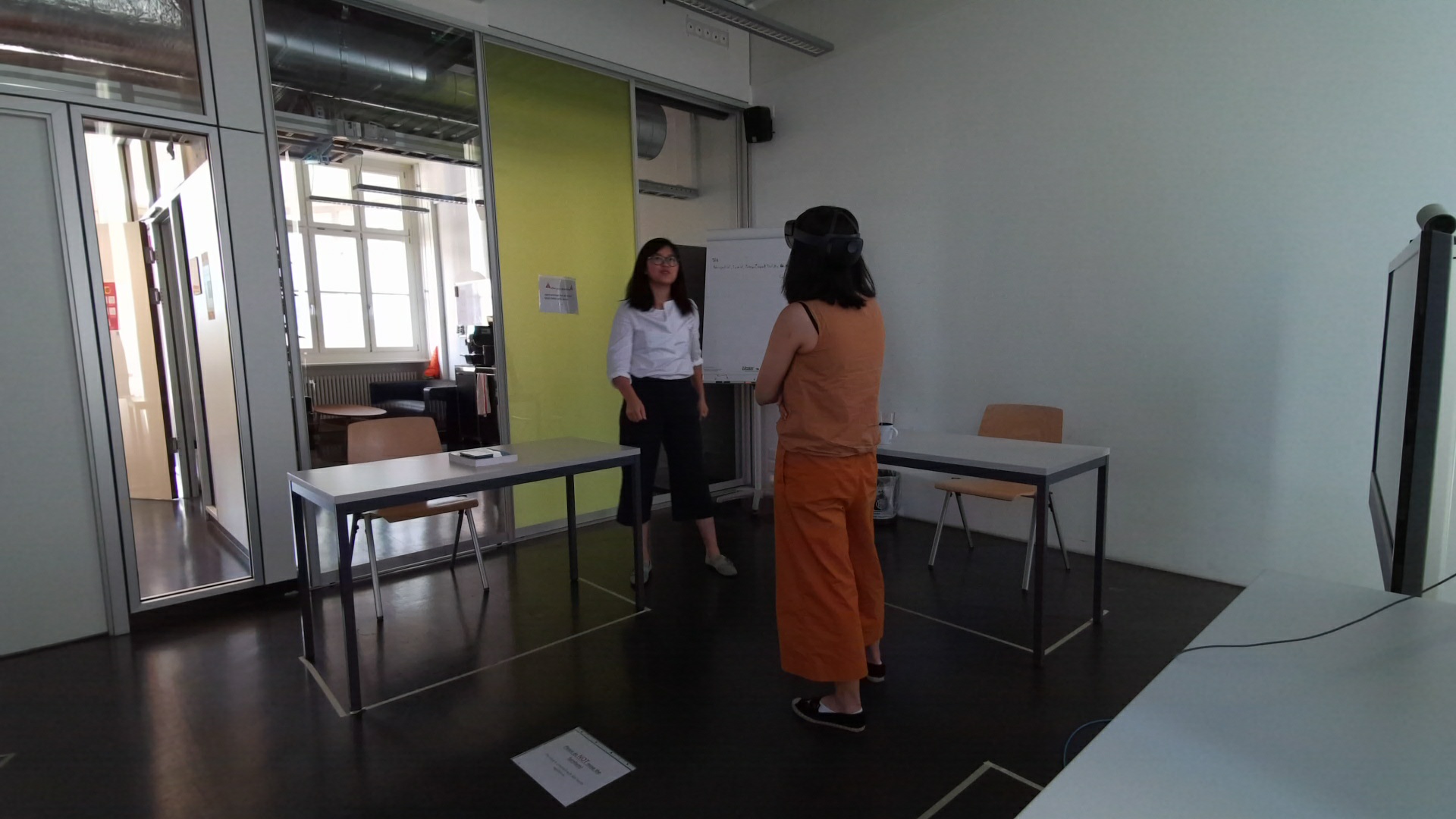}
    \includegraphics[width=0.49\linewidth]{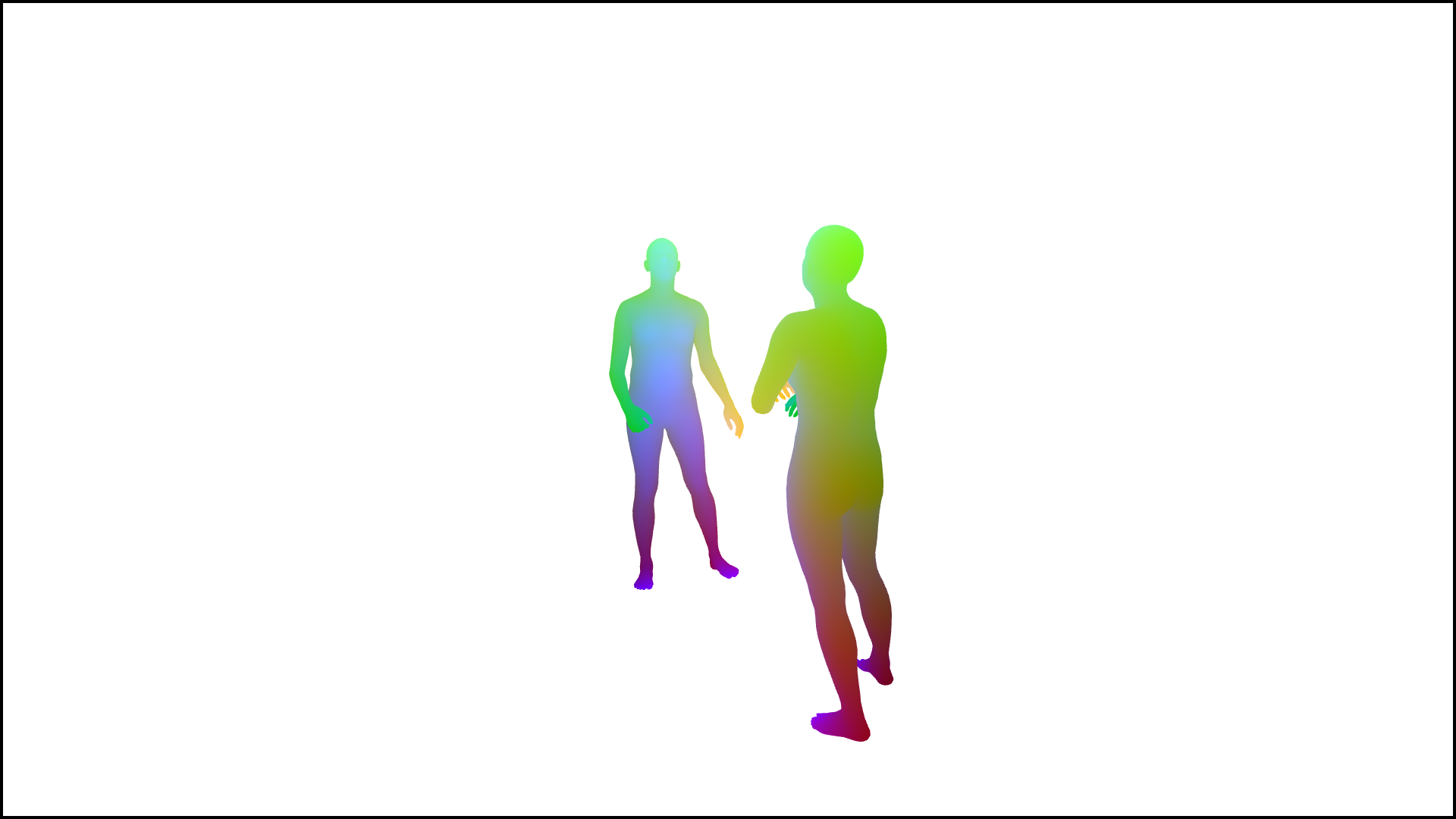} \\[-0.2cm]
    \caption{\textbf{Illustration of the human-centric datasets} used in this paper and listed in Table~\ref{tab:training_datasets}, namely HumGen3D, BEDLAM, HuMMan and EgoBody. For each dataset, we show an input image (left), along with its corresponding DensePose annotations (right).}
    \label{fig:data}
    \vspace{-0.4cm}
\end{figure}

\paragraph{Training Datasets.} Obtaining large-scale, multi-view data with accurate camera poses, depth maps, and parametric body annotations is particularly challenging for human-centric scenes. Real-world captures often require specialized equipment or extensive manual post-processing, making them both costly and error-prone. Consequently, we rely primarily on synthetic data where camera intrinsics, poses, and depth can be automatically recorded during rendering.
We generate our own training data with the following pipeline: For each person in the data, we first sample a random body shape and pose from the AMASS dataset~\cite{amass}, then map a human model to it, from the \textit{HumGen3D}~\cite{humgen3d} human generator plugin for Blender. These Humgen3D humans are then placed in detailed indoor 3D scenes that are procedurally generated with Infinigen Indoors~\cite{infinigen}. Finally, these scenes are rendered with Blender, along with the necessary annotations (depth, instance masks, and DensePose). We generate 524k images, rendered from 10k scenes and 1000 unique 3D environments, with approximately 5 persons per scene.
To increase diversity in environment types and the number of subjects, we also incorporate BEDLAM~\cite{bedlam}, which includes both indoor and outdoor settings, and HuMMAN~\cite{humman}, which features single individuals performing complex poses. Lastly, we include EgoBody~\cite{egobody}, a real dataset captured with multiple Kinect sensors that contains up to two individuals per scene and provides accurate depth maps. A summary of these datasets is provided in \Table{training_datasets} and some examples are shown in Figure~\ref{fig:data}.

%% file: tables/datasets.tex
\begin{table}[t]
\centering
\caption{\textbf{Human-centric training datasets} used for finetuning on human-centric scenes alongside the original \master training datasets. All datasets provide camera pose, depth, and instance segmentation. BEDLAM's 9.2k scenes are motion clips from 8 3D environments and 95 HDRIs. For EgoBody, we use an off-the-shelf segmentation tool~\cite{gsam} to obtain the instance segmentations. 
}
\vspace{-0.2cm}
\label{tab:training_datasets}
\resizebox{\columnwidth}{!}{%
    \begin{tabular}{lccc}
        \toprule
        \textbf{Dataset} & \textbf{Domain} & \textbf{Type} &  \textbf{Scenes} \\
        \midrule
        HumGen3D      & Synthetic & Indoor  & 10k     \\
        BEDLAM~\cite{bedlam}         & Synthetic & Indoor \& Outdoor      & 9.2k    \\
        HuMMan~\cite{humman} & Real    & Studio              & 339     \\
        EgoBody~\cite{egobody} & Real    & Indoor             & 125      \\
        \bottomrule
    \end{tabular}
    }
    \vspace{-0.4cm}
\end{table}

%% file: sec/4_exps.tex
\input{tables/hmr} 
\section{Experimental results}
\label{sec:exp}

We evaluate our approach on both human-centric and traditional 3D tasks.
We first present the evaluation protocol in Section~\ref{sub:xpeval} and then discuss results in Section~\ref{sub:xpres}.

\subsection{Datasets and Metrics}
\label{sub:xpeval}

\paragraph{Human-centric experiments.} We evaluate the effectiveness of \ourmethod across diverse indoor and outdoor environments, covering various activities involving one or multiple individuals.
Following~\cite{hsfm}, we evaluate on EgoHumans~\cite{egohumans} and EgoExo4D~\cite{egoexo},  and report different Mean Per-Joint Position Error (MPJPE) metrics (expressed in meters): W-MPJPE, when measured in the world coordinate system, PA-MPJPE, its Procrustes-Aligned version, and Group-Aligned MPJPE, (GA-MPJPE),  after alignment between people.  To obtain 3D joint predictions with our method, we fit SMPL to our predictions using an optimization procedure that minimizes the distance between all predicted 3D points of the person, and the corresponding vertices on the SMPL model. This is accompanied by an additional loss that serves as a prior on the pose and shape, using VPoser~\cite{smplx} for pose regularization. We build upon the MvSMPLFitting framework~\cite{mvsmplfitting}, which extends SMPLify-X~\cite{smplx} to multi-view settings.
For  evaluating camera poses, we also report  the average camera translation error TE, and its Sim(3) aligned version (s-TE), the camera Angle Error (AE), the Relative Rotation Accuracy (RRA), the Camera Center Accuracy (CCA) and its version computed after Sim(3) alignment (s-CCA).

\paragraph{Traditional 3D tasks.} To ensure that our model still performs well in classical 3D vision tasks, we also evaluate it for the task of multi-view stereo depth estimation on  KITTI~\cite{kitti}, ScanNet~\cite{scannet}, ETH3D~\cite{eth3d}, DTU~\cite{dtu}, Tanks and Temples~\cite{tankstemples}, following~\cite{dust3r}. We report the Absolute Relative Error (rel) and Inlier Ratio ($\tau$) with a threshold of 1.03 on each test set and the averages across all test sets.
Additionally, we assess the ability of \ourmethod to perform multi-view pose estimation on the CO3Dv2~\cite{co3d} and RealEstate10K~\cite{realestate10K} datasets following~\cite{mast3r} and report the Relative Rotation Accuracy (RRA) and Relative Translation Accuracy (RTA) for each image pair to evaluate the relative pose error and we select a threshold of 15\textdegree~to report $RTA@15$ and $RRA@15$. Additionally, we calculate the mean Average Accuracy (mAA30).

\input{tables/camera_humans}

\subsection{Results}
\label{sub:xpres}

\paragraph{Human-centric experiments.}

We report human metrics in Table~\ref{tab:human_metrics}, comparing various human pose estimation baselines on EgoHumans and EgoExo4D. 
Specifically,
we evaluate  against
UnCaliPose~\cite{uncalipose} and the concurrent work HSfM~\cite{hsfm}, as well as a monocular baseline Multi-HMR~\cite{multihmr}.
For the latter, we simply select a random view among the set, to be used as input.
For HSfM, we also provide numerical evaluation before their optimization step (init).
While both HSfM and UnCaliPose  jointly reconstruct humans and cameras, HSfM is more comparable to our approach as it also reconstructs the environment and leverages DUSt3R to estimate the cameras. \ourmethod outperforms the other baselines on EgoExo4D in World coordinate metric (W-MPJPE = 0.51~m), in particular HSfM (0.56~m) that uses a global scene optimization (and bundle adjustment guided by 2D human keypoint predictions) to optimize the humans, depth maps, and cameras. However, after Procrustes alignment, our performance (PA-MPJPE= 0.09m) is slightly below HSfM’s performance (PA-MPJPE= 0.06m). The similar results of HSfM (init) for  PA-MPJPE (0.07m) indicates that most of its human pose estimation accuracy comes from its strong initialization using the off-the-shelf HMR2~\cite{hmr2}.  
A possible reason for our slightly lower performance in PA-MPJPE is that we discard RGB information before fitting the SMPL model, making it challenging to match neural HMR methods that leverage richer appearance cues. 
The same trend is observed on EgoHumans, where we outperform the state-of-the-art method UnCalibPose on W-MPJPE and GA-MPJPE but not on PA-MPJPE. Notably, on this dataset, the concurrent work HSfM achieves significantly better results, particularly on the PA-MPJPE metric (0.17m for us \vs 0.05m for HSfM), where their global optimization provides only marginal improvements over their strong initialization (0.06m). Our lower performance on EgoHumans compared to EgoExo4D is likely due to the nature of scenes, which often features  large, open, outdoor environments.
Our method to estimate SMPL parameters appears more sensitive to lower-resolution inputs as obtaining accurate SMPL fits becomes difficult when a person occupies only a small portion of the image.

\input{tables/4_cam_average}
The camera pose estimation metrics are reported in Table~\ref{tab:camera_metrics}, where we evaluate our method against HSfM and UnCaliPose, as well as the original DUSt3R and MASt3R. To better analyze our performance, we also report camera metrics when estimating the cameras using \divine.
On EgoExo4D,  \ourmethod clearly outperforms all the other baselines for all the considered metrics. On the EgoHumans dataset, \divine~consistently achieves the best performance overall. In contrast, training our method seems to reduce the accuracy of camera estimates. We hypothesize that this is due to the large scale of some environments in the EgoHumans dataset, combined with our use of \master log-scaling of the 3D regression loss.
This non-linear scaling does not penalize distant points as heavily, which may negatively affect performance on larger scenes. This hypothesis is further supported by separate evaluations of performance on large and small scenes, where our method performs significantly better on smaller environments; See experiments in Table~\ref{tab:camera_metrics_scale}.

\input{tables/camera_co3d}

\input{tables/reconstruction}

\begin{figure*}[ht]
    \centering
    
    \includegraphics[width=\textwidth]{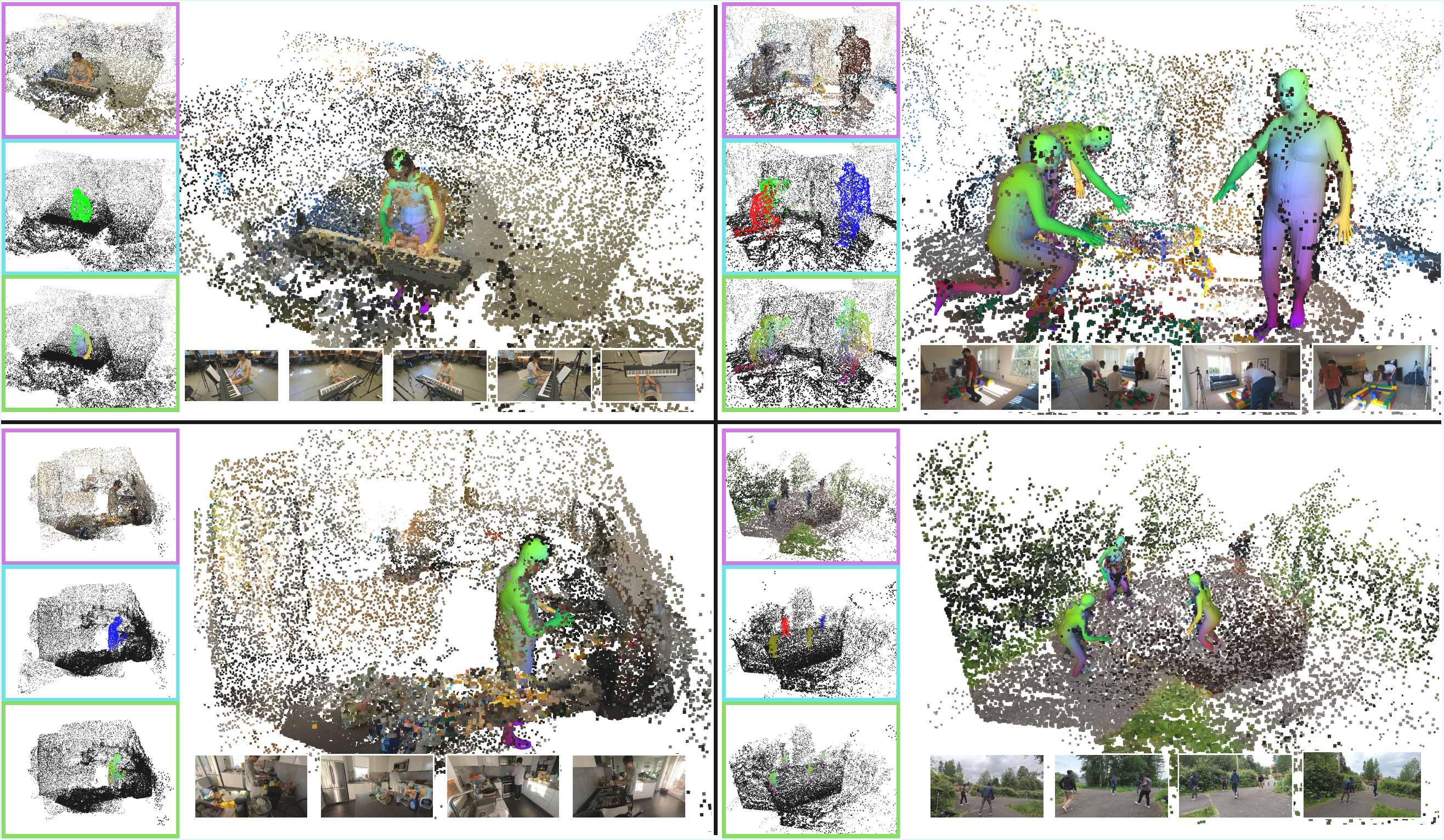} \\[-0.3cm]
    \definecolor{pastelpink}{HTML}{CB6CE6}  
    \definecolor{pastelblue}{HTML}{5CE1E6}
    \definecolor{pastelgreen}{HTML}{7ED957}

    \caption{\textbf{Qualitative results of \ourmethod.} Results from EgoExo4D (left) and EgoHumans (right). Each example includes point clouds (\textcolor{pastelpink}{pink}), instance segmentation (\textcolor{pastelblue}{blue}), and dense pose (\textcolor{pastelgreen}{green}), with corresponding input images below (Best viewed when zoomed in). }
    \label{fig:qualitative_results}
    \vspace{-0.3cm}
\end{figure*}

\paragraph{Traditional 3D tasks.} 
Table~\ref{tab:multiview_depth} presents our multi-view stereo depth estimation results. As expected, overall performance decreases compared to the original DUSt3R and \master models. This is due to two factors: (1) we use DUNE's encoder that distills MASt3R alongside an HMR model, enhancing human understanding but reducing depth accuracy for structures (see the drop in performance between \master and \divine) and buildings, and (2) we introduce additional human-centric heads and tasks during training while using 50\% human-centric data, further impacting depth estimation in non-human regions. Nevertheless, \ourmethod remains competitive, performing on par with or even surpassing recent deep learning architectures such as Deepv2D~\cite{deepv2d}. A similar trend is observed in the multi-view pose regression results reported in Table~\ref{tab:multiview_pose}, where performance drops on CO3Dv2 but unexpectedly improves on RealEstate10K. The distillation process is beneficial on this benchmark (see \divine's performance) and the performance drop is less important when training \ourmethod.


\paragraph{Qualitative results} are presented in Figure~\ref{fig:qualitative_results}, 
illustrating human reconstructions from EgoExo4D and EgoHumans. Our method successfully estimates dense human poses, instance segmentation, and point clouds across a variety of indoor and outdoor settings. The reconstructions capture realistic human shapes and spatial configurations, demonstrating the robustness of our approach even in challenging scenes. While some minor artifacts are visible in complex environments, our results align well with the quantitative findings, reinforcing our strong performance in world-coordinate metrics.

%% file: tables/hmr.tex
\begin{table*}[ht]
  \caption{\textbf{Human-centric evaluation metrics} on EgoHumans and EgoExo4D.}
  \vspace{-0.3cm}
  \centering
  \resizebox{13cm}{!}{%
    \begin{tabular}{l|ccc|cc}
      \toprule
      \multirow{2}{*}{\textbf{Method}} & \multicolumn{3}{c|}{\multirow{1}{*}{\textbf{EgoHumans}}}  & \multicolumn{2}{c}{\multirow{1}{*}{\textbf{EgoExo4D}}} \\
      & W-MPJPE $\downarrow$ & GA-MPJPE $\downarrow$ & PA-MPJPE $\downarrow$ & W-MPJPE $\downarrow$  & PA-MPJPE $\downarrow$ \\
      \midrule
     
       Multi-HMR~\cite{multihmr}           & 7.66 & 0.99 & 0.12  & 2.88 & 0.07 \\
       UnCaliPose~\cite{uncalipose} & 3.51 & 0.67 & 0.13 & 2.90 &  0.13 \\
              HSfM (init)~\cite{hsfm}           & 4.28 & 0.51  & 0.06  & 5.29 &  0.07 \\
       HSfM~\cite{hsfm}           & {\bf 1.04} & {\bf 0.21} & {\bf 0.05}  &  0.56 &  {\bf 0.06} \\

       \textbf{{\bf \ourmethod (Ours)}}       & 3.80   & 0.42   & 0.14     & {\bf {0.51}}    &  0.09    \\
      \bottomrule
    \end{tabular}%
  }
  \label{tab:human_metrics}
\end{table*}

%% file: tables/camera_humans.tex
\begin{table*}[ht]
    \centering
    \caption{\textbf{Camera pose evaluation} on EgoHumans and EgoExo4D.}
    \vspace{-0.3cm}
    \resizebox{\textwidth}{!}{%
        \begin{tabular}{l|cccccc|cccccc}
        \toprule
        \multirow{2}{*}{\textbf{Method}} 
            & \multicolumn{6}{c|}{\textbf{EgoHumans}} 
            & \multicolumn{6}{c}{\textbf{EgoExo4D}} \\
        & TE $\downarrow$ & s-TE $\downarrow$ & AE $\downarrow$ & RRA@10 $\uparrow$ & CCA@10 $\uparrow$ & s-CCA@10 $\uparrow$
        & TE $\downarrow$ & s-TE $\downarrow$ & AE $\downarrow$ & RRA@10  $\uparrow$& CCA@10 $\uparrow$ & s-CCA@10 $\uparrow$\\
        \midrule
        
        UnCaliPose~\cite{uncalipose} 
            & 2.63 
            & 2.63 
            & 60.90 
            & 0.28 
            & - 
            & 0.33 
            & 2.43 
            & 1.16 
            & 65.61 
            & 0.19 
            & - 
            & 0.24 \\
            
        DUSt3R~\cite{dust3r} 
            & -    
            & 1.15 
            & 11.00 
            & 0.61 
            & - 
            & 0.49 
            & -    
            & 0.33 
            & 9.92 
            & 0.81 
            & - 
            & 0.64 \\
            
        MASt3R~\cite{mast3r} 
            & 4.97 
            & 0.92 
            & 10.42 
            & 0.61 
            & 0.06 
            & 0.65 
            & 0.96 
            & 0.35 
            & 11.70 
            & 0.79 
            & 0.06 
            & 0.68 \\
        HSfM (init)~\cite{hsfm} 
            & 2.37 
            & 1.15 
            &  11.00 
            & 0.52 
            & 0.26
            & 0.49
            & 1.27 
            & 0.33 
            & 9.92 
            & 0.81
            & 0.05 
            & 0.64 \\
            
        HSfM~\cite{hsfm} 
            & \underline{2.09} 
            & 0.75 
            & \underline{9.35} 
            & 0.72 
            & \textbf{0.32} 
            & \underline{0.75} 
            & \underline{0.95} 
            & 0.36 
            & 11.57 
            & 0.78 
            & 0.07 
            & 0.67 \\
            
        \divine 
            & \textbf{1.43} 
            & \textbf{0.17} 
            & \textbf{3.51} 
            & \textbf{0.96} 
            & \underline{0.27} 
            & \textbf{0.97} 
            & 1.03 
            & \underline{0.26} 
            & \underline{6.45} 
            & \underline{0.94} 
            & \underline{0.25} 
            & \underline{0.85} \\
            
        \textbf{\ourmethod (Ours)} 
            & 2.33 
            & \underline{0.40} 
            & 10.24 
            & \underline{0.77} 
            & 0.06 
            & \underline{0.75} 
            & \textbf{0.60} 
            & \textbf{0.15} 
            & \textbf{2.85} 
            & \textbf{0.99} 
            & \textbf{0.42} 
            & \textbf{0.87} \\
        
        \bottomrule
        \end{tabular}
    }
    \label{tab:camera_metrics}
    \vspace{-0.2cm}
\end{table*}

%% file: tables/4_cam_average.tex
 \begin{table}[t]
    \centering
    \caption{\textbf{Camera pose evaluation on EgoHumans according to scene scale.} Average results over the entire dataset (All), compared with a finer analysis based on the split between large-scale/open scenes (Badminton, Tennis, Volleyball) and smaller environments (Basketball, Fencing, Lego, Tagging).}
    \vspace{-0.3cm}
    \resizebox{\columnwidth}{!}{%
    \begin{tabular}{llcccccc}
     \midrule
    Env. & Method & TE $\downarrow$ & s-TE $\downarrow$ & AE $\downarrow$ & RRA@10 $\uparrow$ & CCA@10 $\uparrow$ & s-CCA@10 $\uparrow$
         \\
        \midrule
        \multirow{2}{*}{All}  
        & \divine  & \textbf{1.43}  & \textbf{0.17}  & \textbf{3.51}  & \textbf{0.96}  & \textbf{0.27}  & \textbf{0.97}   \\
        &   \textbf{\ourmethod (Ours)}    & 2.33  & 0.40  & 10.24 & 0.77  & 0.06  & 0.75   \\
        \midrule
        \multirow{2}{*}{Large}  
         & \divine  & \textbf{0.98}  & \textbf{0.24}  & \textbf{4.78}  & \textbf{0.90}  & \textbf{0.34}  & \textbf{0.93}   \\
        &    \textbf{\ourmethod (Ours)}    & 3.30  & 0.66  & 16.303 & 0.58  & 0.00  & 0.51  \\
       \midrule
        \multirow{2}{*}{Small}  
         & \divine  & 2.04  & \textbf{0.11}  & 2.36  & \textbf{1.00}  & \textbf{0.14}  & 0.99   \\
        &    \textbf{\ourmethod (Ours)}    & \textbf{1.31}  & 0.12  & \textbf{2.32}  & \textbf{1.00}  & \textbf{0.14}  & \textbf{1.00}  \\
        \bottomrule
    \end{tabular}
    }
     \label{tab:camera_metrics_scale}
     \vspace{-0.4cm}
\end{table}

%% file: tables/camera_co3d.tex
\begin{table}[t]
    \centering
    \caption{\textbf{Multi-view pose regression evaluation} on the CO3Dv2~\cite{co3d}  and RealEstate10K~\cite{realestate10K} with 10 random frames.}
    \vspace{-0.3cm}
    \resizebox{\columnwidth}{!}{%
        \begin{tabular}{l|ccc|c}
        \toprule
        \multirow{2}{*}{\textbf{Method}} & \multicolumn{3}{c|}{\textbf{Co3Dv2} $\uparrow$} & \textbf{RealEstate10K} $\uparrow$ \\
        
         & RRA@15 & RTA@15 & mAA(30) & mAA(30) \\
        \midrule
        DUSt3R~\cite{dust3r} & \underline{93.3} & 88.4 & 77.2 & 61.2 \\
        MASt3R~\cite{mast3r} & {\bf 94.6} &  {\bf 91.9} & {\bf 81.8} & 76.4 \\
        \divine              & 92.2 & \underline{90.7} & \underline{78.7} & {\bf 80.1} \\
        \textbf{{\ourmethod~(Ours)}}        & 90.7 & 90.2 & 76.3 & \underline{77.4} \\
        \bottomrule
        \end{tabular}
    }
    \label{tab:multiview_pose}
    \vspace{-0.4cm}
\end{table}

%% file: tables/reconstruction.tex
\begin{table*}[ht]
    \centering
    \caption{\textbf{Multi-view depth evaluation} (ScanNet) denote training on data from the same domain.}
    \vspace{-0.3cm}
    \resizebox{14cm}{!}{%
    \begin{tabular}{l|c@{~~~~}cc@{~~~~}cc@{~~~~}cc@{~~~~}cc@{~~~~}cc@{~~~~}c}
\toprule
\multirow{2}{*}{\textbf{Method}} &  \multicolumn{2}{c}{\textbf{KITTI}}             & \multicolumn{2}{c}{\textbf{ScanNet}}           & \multicolumn{2}{c}{\textbf{ETH3D}}             & \multicolumn{2}{c}{\textbf{DTU}}               & \multicolumn{2}{c}{\textbf{T\&T}}              & \multicolumn{2}{c}{\textbf{Average}}           \\
               & rel. $\downarrow$ & $\tau$ $\uparrow$ & rel. $\downarrow$ & $\tau$ $\uparrow$ & rel. $\downarrow$ & $\tau$ $\uparrow$ & rel. $\downarrow$ & $\tau$ $\uparrow$ & rel. $\downarrow$ & $\tau$ $\uparrow$ & rel. $\downarrow$ & $\tau$ $\uparrow$ \\ \midrule
DeepV2D (ScanNet)~\cite{deepv2d}  & 10.00              & 36.20             & \textbf{4.40}              & 54.80             & 11.80              & 29.30             & 7.70              & 33.00             & 8.90             & 46.40            & 8.60              & 39.90             \\               
DUSt3R~\cite{dust3r}  & ~~5.88              & 47.67             & \textbf{3.01}              & \textbf{72.54}             & ~~3.04              & 75.17             & 2.92              & 73.94             & 2.93              & 78.51             & 3.56              & 69.56             \\
MASt3R~\cite{mast3r}  & ~~\textbf{3.54 }             & \textbf{65.68}            & 4.17              & 65.22             & ~~\textbf{2.44}              & \textbf{82.77}             & 3.46              & 66.89             & \textbf{2.04}              & \textbf{87.88}             & \textbf{3.13}              & \textbf{73.69}             \\
\divine~\cite{dune}   & \underline{4.88}             & 50.76             & 4.24              & 59.68             & ~~2.48              & 77.97             & \textbf{2.69}              & \textbf{75.63}             & 2.60              & 79.19             & 3.38              & 68.65             \\
\textbf{{\ourmethod~(Ours)}}    & 5.60             & 45.66             & 4.43              & 56.50             & ~~2.96              & 71.68             & 5.31              & 57.62             & 3.01              & 73.53             & 4.26              & 61.00             \\ \bottomrule

\end{tabular}
    }
    \label{tab:multiview_depth}
    \vspace{-0.25cm}
\end{table*}

%% file: sec/5_conclusions.tex
\section{Conclusion}

We have introduced \ourmethod, the first feed-forward method for jointly reconstructing people and their surroundings from sparse stereo views. Given multiple images of a scene involving one or several persons, it produces dense point maps with human semantic information in 3D.  Unlike optimization-based approaches, our method avoids common drawbacks such as computational slowness and sensitivity to hyperparameters. 
Through extensive evaluation, we demonstrate that our approach,  combined with SMPL fitting, outperforms prior methods in estimating human poses and achieves competitive results to concurrent work in small environments, all while maintaining strong performance in general scene reconstruction—even in the absence of people.  For future work, we aim to extend our method to videos and dynamic scenes, further enhancing its applicability to real-world scenarios.

%% file: sec/10_supp.tex
\clearpage
\appendix

\section*{Supplementary Material}

\label{sec:supp}
\subsection*{Loss Weight Selection} 
The loss weights \(\lambda_1\), \(\lambda_2\), and \(\lambda_3\) were selected to balance the contributions of the segmentation, DensePose, and binary mask losses, respectively. To ensure all loss terms operate on a comparable numerical scale, we first computed the typical magnitude of each loss on a held-out validation set and then scaled the corresponding \(\lambda\) accordingly. The final values used in our experiments were \(\lambda_1 = 0.01\), \(\lambda_2 = 1\), and \(\lambda_3 = 1\).

\subsection*{Runtime Comparison}
To contextualize the efficiency of our approach, we report end-to-end runtimes on a representative 4-view input of a 3-person scene (EgoExo4D-style setup). All measurements were performed on a single NVIDIA V100 GPU.

\noindent HSfM uses a multi-stage optimization pipeline with external tools, each adding to the total runtime. In contrast, HAMSt3R is a unified, feed-forward architecture with optional SMPL fitting used only for evaluation.

Despite being a unified feed-forward model, HAMSt3R remains highly efficient compared to modular pipelines such as HSfM. SMPL fitting is performed only as a post-processing step for evaluation.

\input{tables/runtime}

\subsection*{Monocular Prediction}
\begin{figure}
\includegraphics[width=0.48\textwidth]{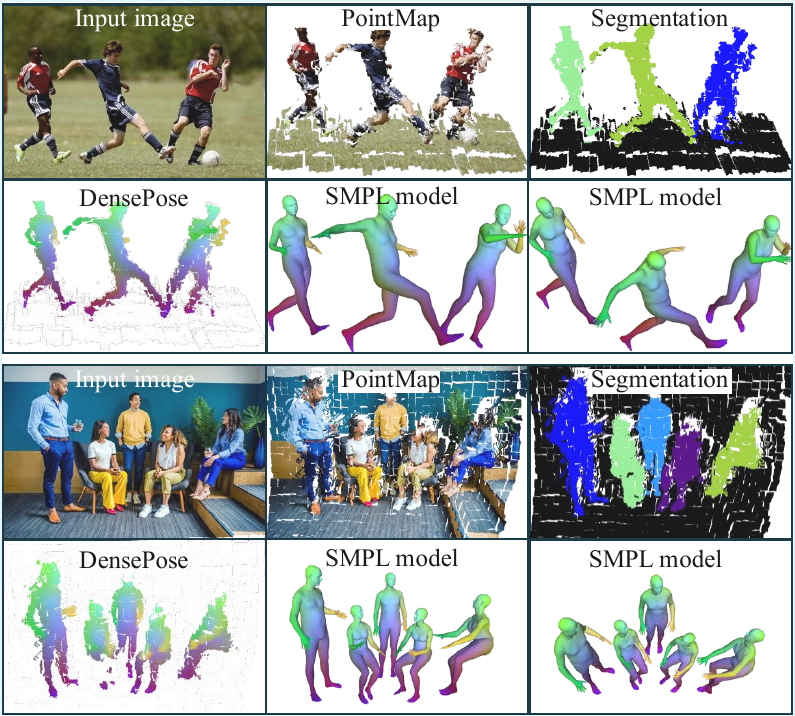}
 \caption{\textbf{Qualitative results of monocular prediction} on in-the-wild images taken from Pexels~\cite{pexels}. Each 2-row group shows: (top row, left to right) input image, high-confidence reconstructed point cloud overlaid with color (PointMap), and segmentation results; (bottom row) DensePose predictions and two different views of the fitted SMPL models. The figure illustrates that our method can produce coherent reconstructions and mesh predictions from a single image.
}
\label{fig:monocularpred}
\end{figure}

HAMSt3R can also be run on a single image, by simply feeding the same image twice to the network. We show some qualitative results on in-the-wild images in Figure~\ref{fig:monocularpred}.

\subsection*{Failure Cases in SMPL Fitting}
\begin{figure}[t]
    \centering
    \includegraphics[width=0.48\textwidth]{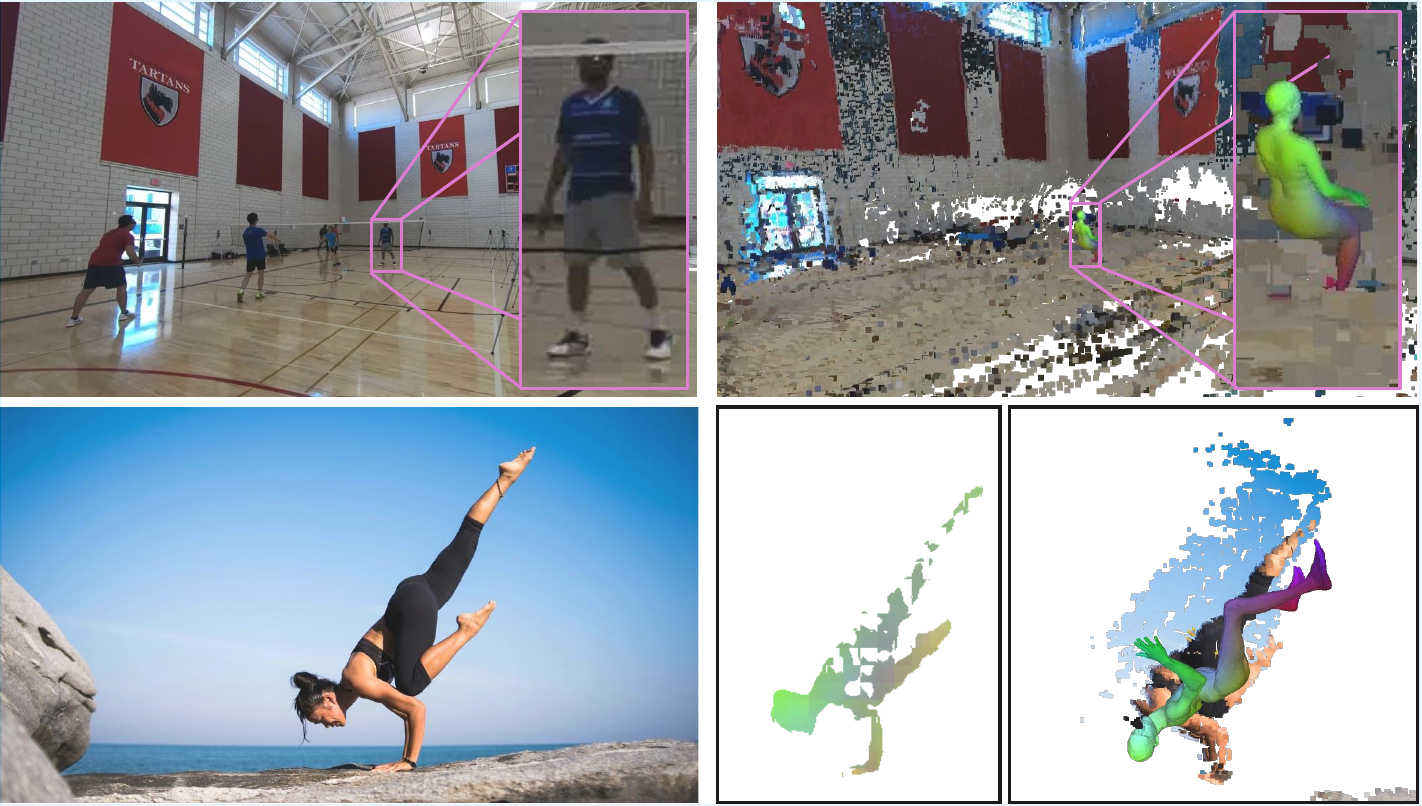}

    \caption{
\textbf{Examples of failure cases.} \textit{Top row}: reconstruction failures on a scene from EgoHumans. Due to the large scene scale and wide camera angle (left), human subjects appear very small after downscaling, leading to noisy point clouds (right) and incorrect orientation in the SMPL mesh (inset). \textit{Bottom row}: failures caused by extreme body poses. DensePose predictions (middle) can break down under uncommon configurations (left), resulting in erroneous SMPL fits (right).
}
    \label{fig:failure}
\end{figure}

While HAMSt3R is generally robust, we observe occasional failures in SMPL fitting, especially when subjects are far from the camera. In such cases, the 3D points are sparse and noisy, leading to unstable optimization (Figure~\ref{fig:failure}).

We hypothesize that increasing reconstruction resolution or incorporating additional priors could improve robustness in these edge cases. These examples highlight the challenges of downstream mesh fitting in low-density areas, even when the upstream reconstruction is geometrically correct.

%% file: tables/runtime.tex
\begin{table}[h]
\centering
\caption{\textbf{Runtime comparison} for HAMSt3R and HSfM in a 4-view, 3-person scene. HAMSt3R is significantly faster despite producing comparable or stronger results.}
\vspace{-0.2cm}
 \resizebox{\columnwidth}{!}{%
\begin{tabular}{l|c}
\toprule
\textbf{Method} & \textbf{Total Runtime (4-view)} \\
\midrule
\textbf{HAMSt3R (Ours)} \\
\quad Reconstruction + Segmentation + DensePose & $\sim$14s \\
\quad SMPL Fitting (post-process) & $\sim$6s \\
\quad \textbf{Total} & $\sim$\textbf{32s} \\
\midrule
\textbf{HSfM~\cite{hsfm}} \\
\quad 2D Pose Initialization & $\sim$1s \\
\quad Segmentation (SAM~\cite{gsam}) & $\sim$2s \\
\quad 3D Reconstruction (DUSt3R~\cite{dust3r}) & $\sim$18s \\
\quad Stage 1 (Translation \& Scale Only) & $\sim$25s \\
\quad Stage 2 (Add Global Orientation \& Align DUSt3R) & $\sim$48s \\
\quad Stage 3 (Add Local Body Pose) & $\sim$24s \\
\quad \textbf{Total} & $\sim$\textbf{118s} \\
\bottomrule
\end{tabular}
}
\label{tab:runtime}
\end{table}